\documentclass[11pt,a4paper]{article}

\usepackage{comment}
\usepackage{graphicx}
\usepackage{amsmath}
\usepackage{amsfonts}
\usepackage{amssymb}
\usepackage{enumerate}
\usepackage{lscape}
\usepackage{longtable}
\usepackage{rotating}
\usepackage{color}
\usepackage{url}
\usepackage{rotating}
\usepackage{algpseudocode}
\usepackage[ruled]{algorithm}
\usepackage{pgf,tikz}
\usepackage{subfig}
\usepackage[titletoc]{appendix}
\usepackage{wrapfig}
\usepackage{adjustbox}
\usepackage{mathrsfs}
\usepackage{pgfplots}
\usetikzlibrary{arrows}
\usepackage{color}

\usepackage{bbm}

\usepackage{booktabs}
\usepackage{multirow}
\usepackage{mathtools}
\usepackage{makecell}
\usepackage{hyperref}

\usepackage[sort,numbers]{natbib}
\usepackage{eqparbox}%

\numberwithin{equation}{section}%

\usepackage[symbol]{footmisc}%

\DeclareMathOperator{\NN}{NN}

\newcommand{\beqn}[1]{\begin{equation}\label{#1}}
\newcommand{\eeqn}{\end{equation}}

\usepackage{accents}
\newlength{\dhatheight}
\newcommand{\doublehat}[1]{%
    \settoheight{\dhatheight}{\ensuremath{\hat{#1}}}%
    \addtolength{\dhatheight}{-0.35ex}%
    \hat{\vphantom{\rule{1pt}{\dhatheight}}%
    \smash{\hat{#1}}}}

\definecolor{darkgreen}{rgb}{0,0.6,0}
\definecolor{aau2}{rgb}{0.0, 0.5, 0.69}
\definecolor{aau3}{rgb}{0.0, 0.53, 0.74}
\definecolor{aau4}{rgb}{0.0, 0.48, 0.65}
\definecolor{aau5}{rgb}{0.0, 0.45, 0.73}
\definecolor{rsap}{RGB}{130, 36, 51}
\definecolor{gsap}{RGB}{112, 164, 137}
\definecolor{tud}{rgb}{0.43,0.73,0.11}
\definecolor{verde}{rgb}{0.33,0.53,0.11}

\definecolor{ttffqq}{rgb}{0.0, 0.48, 0.65} %{rgb}{0.43,0.73,0.11}
\definecolor{ffqqqq}{rgb}{0.0, 0.5, 0.69} %{rgb}{1,0,0}
\usetikzlibrary{arrows}

\tikzstyle{decision} = [diamond, draw, fill=blue!20,
text width=4.5em, text badly centered, node distance=3cm, inner sep=0pt]
\tikzstyle{block} = [rectangle, draw, fill=blue!20,
text centered, rounded corners, minimum height=4em]
\tikzstyle{line} = [draw, -latex']
\tikzstyle{cloud} = [draw, ellipse,fill=red!20, node distance=3cm,
minimum height=2em]
\tikzstyle{cloud2} = [draw, ellipse,fill=green!20, node distance=3cm,
minimum height=2em]
\begin{document}
	
	\title{The limitation of neural nets for approximation and optimization}
	
	\author{
		T. Giovannelli\thanks{Department of Industrial and Systems Engineering, Lehigh University, Bethlehem, PA 18015-1582, USA ({\tt tog220@lehigh.edu}).}
		\and
		O. Sohab\thanks{Department of Industrial and Systems Engineering, Lehigh University, Bethlehem, PA 18015-1582, USA ({\tt ous219@lehigh.edu}).}
		\and
		L. N. Vicente\thanks{Department of Industrial and Systems Engineering, Lehigh University, Bethlehem, PA 18015-1582, USA ({\tt lnv@lehigh.edu}).}
	}
% Support for this author was partially provided by the Centre for Mathematics of the University of Coimbra under grant FCT/MCTES UIDB/MAT/00324/2020.	
	\maketitle
	
	\begin{abstract}
We are interested in assessing the use of neural networks as surrogate models to approximate and minimize objective functions in optimization problems. While neural networks are widely used for machine learning tasks such as classification and regression, their application in solving optimization problems has been limited. Our study begins by determining the best activation function for approximating the objective functions of popular nonlinear optimization test problems, and the evidence provided shows that~SiLU has the best performance. We then analyze the accuracy of function value, gradient, and Hessian approximations for such objective functions obtained through interpolation/regression models and neural networks. When compared to interpolation/regression models, neural networks can deliver competitive zero- and first-order approximations (at a high training cost) but underperform on second-order approximation. 
However, it is shown that combining a neural net activation function with the natural basis for quadratic interpolation/regression can waive the necessity of including cross terms in the natural basis, leading to models with fewer parameters to determine. 
Lastly, we provide evidence that the performance of a state-of-the-art derivative-free optimization algorithm can hardly be improved when the gradient of an objective function is approximated using any of the surrogate models considered, including neural networks.
	\end{abstract}
 
%%%%%%%%%%%%%%%%%%%%%%%%%%%%%%%%%%%%%%%%%%%%%%%%%%%%%%%%%%%%%%%%%%%%%%%%%%%%%%%%%%%%%%%%%
	\section{Introduction}
%%%%%%%%%%%%%%%%%%%%%%%%%%%%%%%%%%%%%%%%%%%%%%%%%%%%%%%%%%%%%%%%%%%%%%%%%%%%%%%%%%%%%%%%%

 In recent years, machine learning (ML) models have been used to enhance optimization algorithms with the goal of improving their performance. One popular approach is to use ML models to approximate the objective function being minimized. These models, called surrogate models, can learn from past evaluations of the objective function and predict its value for new inputs (see the review in~\cite{JVSoaresDoAmaral_etal_2022}). Using surrogate models can be particularly useful for optimization problems where each evaluation of the objective function is time-consuming or computationally expensive, like in simulation-based optimization or derivative-free optimization~(DFO)~\cite{ARConn_KScheinberg_LNVicente_2009,CAudet_WHare_2017,JLarson_MMenickelly_SMWild_2019,ALCustodio_KScheinberg_LNVicente_2017}. Once an accurate and computationally cheap surrogate model is built, one can evaluate such a model instead of the true objective function to reduce the number of objective function evaluations, thus improving optimization efficiency. Another approach is to use ML models to learn the optimization process itself. These models can iteratively learn from past optimization runs and predict the best algorithmic steps and hyperparameters of methods applied to solve a given problem. For example, in~\cite{YRuan_etal_2019} and~\cite{YChen_MWHoffman_2016}, the authors train recurrent neural networks to learn the update steps of gradient-based and derivative-free optimization algorithms. Such networks are then used to predict the next point to evaluate without adding to the number of objective function and gradient evaluations.

 For the scope of this paper, among all the different types of~ML models, we focus on artificial neural networks. Such networks consist of interconnected nodes (also called neurons) that are organized into layers, typically consisting of an input layer, one or more hidden layers, and an output layer~\cite{IGoodfellow_etal_2016}. The inputs to the network are fed into the input layer, and then they are passed through the hidden layers, where each neuron receives inputs from the previous layer, computes a weighted sum of inputs, and then outputs the result to the next layer after applying an activation function. The output layer produces the final output of the network. Among all the different types of artificial neural networks, we consider feedforward neural networks, where the inputs are processed in a forward direction only, without any feedback connections.

The ultimate goal of this paper is to assess the use of neural networks as surrogate models for approximation and optimization purposes. Neural networks have gained significant popularity as surrogate models in engineering applications~\cite{IPan_etal_2014,VPapadopoulos_etal_2018,MCMessner_2019,KSlimani_MZaaf_TBalan_2023}. However, we are not aware of any papers evaluating the accuracy of the approximations produced by neural networks or using neural networks within optimization algorithms to approximate the objective function being minimized. For assessing the performance of neural networks, we choose to compare them against models built through interpolation or regression, including quadratic models~\cite{ABhaduri_etal_2020,MZaborski_JMandziuk_2023,ARConn_PhLToint_1996,ARConn_KScheinberg_LNVicente_2005,ARConn_KScheinberg_LNVicente_2006,ARConn_KScheinberg_LNVicente_2009,ASBandeira_KScheinberg_LNVicente_2012} and interpolation models based on radial basis functions~\cite{MDBuhmann_2003,DBMcDonald_WJGrantham_2007,HMGutmann_2001}, which have successfully been used for approximation and optimization. To conduct our study, we consider popular nonlinear optimization test problems with different features in terms of linearity, convexity, and separability. It is important to note that our focus in this paper is both on the local and global behavior of the approximations provided by the surrogate models considered. In particular, we are interested in the accuracy of gradient and Hessian approximations for a given point in the domain and in the accuracy of function value approximations across a larger set of points.

The main contributions of this paper can be summarized as follows: 
\begin{enumerate}
    \item 
Determine the best activation function for approximating the objective functions of popular nonlinear optimization test problems. The evidence provided shows that~SiLU has the best performance (see Section~\ref{sec:activation_functions}). 
\item Analyze the accuracy of the function value, gradient, and Hessian approximations for the objective functions in the test problems obtained using interpolation/regression models, including quadratic models and interpolation models based on radial basis functions, and neural networks.  The evidence provided shows that the interpolation/regression models provide better accuracy when they are used for second-order approximations. Neural networks are competitive in terms of function evaluations required for zero- and first-order approximations (although at a high training cost) but perform below interpolation/regression models in terms of second-order approximations. Interestingly, the results also show that combining an activation function with the natural basis for quadratic interpolation/regression can waive the necessity of including cross terms in the natural basis, leading to models with fewer parameters to determine (see Section~\ref{sec:basis_functions}).
\item Present evidence that the performance of a state-of-the-art DFO algorithm can hardly be improved when the gradient of the objective functions in the test problems is approximated using any of the surrogate models considered, including neural networks (see Section~\ref{sec:optimizing_without_derivatives}).
When doing finite-difference BFGS, it seems that no surrogate modeling helps, including a magical/search step based on surrogate optimization.
\end{enumerate}

We focus on unconstrained optimization problems with a continuously differentiable objective function $f: \mathbb{R}^{n} \rightarrow \mathbb{R}$, namely,
\begin{equation}\label{prob:DFO}
    \min_{x \in \mathbb{R}^{n}} f(x).
\end{equation}
The derivatives of~$f$ will be computed for assessing the accuracy of the approximations provided by surrogate models. However, they will be assumed unavailable when solving problem~\eqref{prob:DFO} in Section~\ref{sec:optimizing_without_derivatives}. 

All the computational experiments described in this paper were run on a Linux server with~32~GB of~RAM and an~AMD Opteron~6128 processor running at~2.00~GHz. For our implementation, we used the PyTorch library available in Python~\cite{APaszke_SGross_etal_pytorch} and the code is publicly available on GitHub.\footnote{\url{https://github.com/sohaboumaima/BasesNNApproxForOpt.git}} Throughout this document, $k$ will be the index used to denote the iterations and $\|\cdot\|$ will be the Euclidean norm. Moreover, given a positive scalar~$\Delta$, $B(x;\Delta) = \{y \in \mathbb{R}^n : \| y - x \| \le \Delta\}$ will denote a closed ball in~$\mathbb{R}^n$ of radius~$\Delta > 0$. 

%%%%%%%%%%%%%%%%%%%%%%%%%%%%%%%%%%%%%%%%%%%%%%%%%%%%%%%%%%%%%%%%%%%%%%%%%%%%%%%%%%%%%%%%%
\section{Best activation function for approximation} \label{sec:activation_functions}
%%%%%%%%%%%%%%%%%%%%%%%%%%%%%%%%%%%%%%%%%%%%%%%%%%%%%%%%%%%%%%%%%%%%%%%%%%%%%%%%%%%%%%%%%

In ML applications, neural networks are widely used for function approximation tasks, such as regression and classification. Activation functions play a critical role in the performance of a neural network model, affecting its ability to approximate functions appearing in complex models. In this section, we provide a review of popular activation functions in ML (see Subsection~\ref{subsec:activation_functions}) and determine the best activation function for approximating the objective functions of popular nonlinear optimization test problems (see Subsection~\ref{subsec:activation_functions_numerical_exp}). 

%%%%%%%%%%%%%%%%%%%%%%%%%%%%%%%%%%%%%%%%%%%%%%%%%%%%%%%%%%%%%%%%%%%%%%%%%%%%%%%%%%%%%%%%%
\subsection{Review of popular activation functions in machine learning}\label{subsec:activation_functions}
%%%%%%%%%%%%%%%%%%%%%%%%%%%%%%%%%%%%%%%%%%%%%%%%%%%%%%%%%%%%%%%%%%%%%%%%%%%%%%%%%%%%%%%%%

 Let us denote the activation function used in a neuron of a neural network as~$s$: $\mathbb{R} \to \mathbb{R}$. Although linear activation functions can be considered (i.e., $s(z) = z$), the activation functions used in practice apply a nonlinear transformation to the inputs of the corresponding neurons. ReLU~\cite{KJarrett_2009}, ELU~\cite{DClevert_TUnterthiner_2015}, SiLU~\cite{SElfwing_EUchibe_KDoya_2017}, Sigmoid~\cite{FRosenblatt_1958,FJRichards_1959,DERumelhart_1986}, and Tanh~\cite{BLKalman_SCKwasny_1992} are some of the most commonly used activation functions in neural networks, and their properties are reviewed below. 
 
ReLU is defined by $s(z) = \max\{0,z\}$ and is the recommended activation function for modern neural networks~\cite[Chapter~6]{IGoodfellow_etal_2016}. Although ReLU is not differentiable at $z=0$, this is usually not problematic in practice since it is unlikely that the derivative of $g$ needs to be precisely evaluated at $z=0$ because of numerical errors. Even if such a derivative needs to be evaluated at $z=0$, software implementations usually provide the left derivative (which is $0$) or the right derivative (which is $1$) without raising any issues~\cite[Chapter~6]{IGoodfellow_etal_2016}. Smooth approximations of~ReLU include~ELU, defined by~$s(z) = z$ for $z > 0$ and $s(z) = \alpha (e^{z} - 1)$ for $z \le 0$, where $\alpha > 0$, and SiLU, defined by $s(z) = z/(1 + e^{-z})$. Another smooth variant is  Softplus~\cite{CDugas_etal_2000,HZheng_ZYang_WLiu_JLiang_YLi_2015}, which is given by $s(z) = \log(1 + e^z)$. However, although Softplus is differentiable everywhere, its use is discouraged because ReLU is still more likely to lead to better results~\cite{XGlorot_etal_2011}. 
 
 The main drawback of ReLU is that it can suffer from the ``dying ReLU'' problem, which occurs when neurons output 0 because of their negative inputs and become permanently inactive during training~\cite{LLu_2020}. To overcome such an issue, \cite{ALMaas_etal_2013} proposed LeakyReLU, which is defined by $s(x) = \max\{0.01z,z\}$. The non-negative slope of LeakyReLU for all $z < 0$ allows neurons to contribute to the network's output even if their inputs are negative. In practice, such an activation function performs comparably to ReLU and, occasionally, it may perform better. An improvement to LeakyReLU is Parametric ReLU~\cite{KHe_etal_2015,BXu_NWang_etal_2015}, where the slope for $z < 0$ is a parameter that needs to be learned while training a neural network. ELU and~SiLU are known to be robust to the dying ReLU issue~\cite{IJahan_MdFAhmed_MdOAli_etal_2023}.

Sigmoid and Tanh have been used extensively in the past but have become largely replaced by ReLU and its variants. Sigmoid, defined by $s(z) = 1/(1 + e^{-z})$, maps each input to $[0,1]$ and is often used to approximate probabilities. Tanh, defined by $s(z) = \frac{e^z - e^{-z}}{e^z + e^{-z}}$, is a rescaled version of Sigmoid that maps input values to~$[-1, 1]$. By following a common convention~\cite{IGoodfellow_etal_2016}, we will refer to Sigmoid and Tanh as the sigmoidal activation functions because they are both s-shaped. The main limitation of the sigmoidal activation functions is that they exhibit slow convergence or get stuck at points that are far from optimality. The reason for this behavior is that both Sigmoid and Tanh become flat when $z$ is very positive or very negative, which leads to derivatives that are close to 0, preventing gradient-based algorithms from making good progress at each iteration (in the ML literature, this issue is referred to as the vanishing gradient problem or the saturation problem).  While Tanh can perform better than Sigmoid in some cases, it still suffers from the vanishing gradient problem for large (positive or negative) input values. For this reason, the use of Sigmoid and Tanh as activation functions is discouraged~\cite[Chapter~6]{IGoodfellow_etal_2016}. As opposed to the sigmoidal activation functions, ReLU and its variants have derivatives equal to 1 for $z > 0$, and this allows gradient-based algorithms to make significant progress in the minimization of the objective function at each iteration.  

Recently, some papers have explored combinations of activation functions or have proposed new activation functions that try to replicate the best properties of ReLU and its variants, Sigmoid, and Tanh. In~\cite{FManessi_ARozza_2018}, two approaches to automatically learn the best linear combination of activation functions during the training phase are proposed. In~\cite{BYuen_etal_2021}, the authors propose a differentiable universal activation function with 5 trainable parameters that allow it to approximate the most commonly used activation functions in ML. In~\cite{GBingham_RMiikkulainen_2020}, activation functions are represented as nodes of a computation graph, and evolutionary search and gradient descent are then used to determine the general form of the best activation function and its parameters, respectively. One of the potential drawbacks of using activation functions with trainable parameters is that it can result in overfitting the training data, leading to poor performance on unseen data. Another potential downside is the extra computational cost of training the model. Since the additional parameters need to be learned during training, the computational complexity and time required for training the model can increase~\cite{KHe_etal_2015,SScardapane_etal_2016}. Therefore, in the numerical experiments reported in Subsection~\ref{subsec:activation_functions_numerical_exp} below, we do not consider activation functions with trainable parameters and we only focus on ReLU, ELU, SiLU, Sigmoid, and Tanh.

%%%%%%%%%%%%%%%%%%%%%%%%%%%%%%%%%%%%%%%%%%%%%%%%%%%%%%%%%%%%%%%%%%%%%%%%%%%%%%%%%%%%%%%%%
% \subsection{Numerical experiments}\label{subsec:activation_functions}
%%%%%%%%%%%%%%%%%%%%%%%%%%%%%%%%%%%%%%%%%%%%%%%%%%%%%%%%%%%%%%%%%%%%%%%%%%%%%%%%%%%%%%%%%

%%%%%%%%%%%%%%%%%%%%%%%%%%%%%%%%%%%%%%%%%%%%%%%%%%%%%%%%%%%%%%%%%%%%%%%%%%%%%%%%%%%%%%%%%
\subsection{Numerical experiments}\label{subsec:activation_functions_numerical_exp}
%%%%%%%%%%%%%%%%%%%%%%%%%%%%%%%%%%%%%%%%%%%%%%%%%%%%%%%%%%%%%%%%%%%%%%%%%%%%%%%%%%%%%%%%%

We conducted experiments to compare the performance of~ReLU, ELU, SiLU, Sigmoid, and~Tanh in the approximation of test functions from two sets of nonlinear unconstrained optimization problems available in the CUTEst library~\cite{NGould_DOrban_PToint_2015}. A first set is composed of~38 problems with user-defined dimensions (in all the experiments reported in this paper, we considered~$n \in \{20,40,60\}$), and a second set is composed of~53 functions with dimensions ranging from 2 to 50. The first set includes problems with different features in terms of non-linearity, non-convexity, and partial separability (see Table~\ref{Tab:40probs} in Appendix~\ref{app:tables} for the names of such problems). The second set was selected from~\cite{SGratton_CRoyer_LNVicente_2018} and contains problems for which negative curvature was detected by running one of the second-order methods reported in~\cite{CPAvelino_JMoguerza_etal_2011} (see Table~\ref{Tab:negcurv_probs} in Appendix~\ref{app:tables} for the names and corresponding dimensions of such problems). 

To determine the best activation function for approximation, we used a neural network as a surrogate model for the objective function of each test problem. The weights of the neural network are denoted as~$w \in \mathbb{R}^{n_w}$. For each test problem, we trained the neural network to minimize the mean squared error between the objective function~$f(x)$ and the surrogate function~$f_{\NN}(x; w)$ over the points~$x^i$ in a training dataset~$\mathcal{D} = \{(x^i,f(x^i)) ~|~i \in \{0,\ldots,N\}\}$. Therefore, we solved the following training problem
\begin{equation}\label{prob:testprob}
    \min_{w \in \mathbb{R}^{n_w}} \mathcal{L}(w; \mathcal{D}) = \sum_{i=0}^{N} ( f(x^i) - f_{\NN}(x^i; w) )^2,
\end{equation}
where~$\mathcal{L}(w; \mathcal{D})$ denotes the empirical risk function over the dataset~$\mathcal{D}$. Problem~\eqref{prob:testprob} was solved five times for each test problem to assess the performance of~ReLU, ELU, SiLU, Sigmoid, and Tanh as activation functions of the neural network used as a surrogate model. In addition to the training dataset~$\mathcal{D}$, we also considered a testing dataset to assess the performance of the surrogate model on data that was not used for training.

To aggregate the results over all the test problems, we used performance profiles~\cite{EDDolan_JJMore_2002,JJMore_SMWild_2007}, which are briefly reviewed in this paragraph. Given a set of solvers~$\mathcal{S}$ and a set of problems~$\mathcal{P}$, let~$t_{p,s} > 0$ be the performance measure of the solver~$s \in \mathcal{S}$ when applied to solve the problem~$p \in \mathcal{P}$. For each solver, we can compute the performance profile~$\rho_s(\alpha)$ as follows
\[
\rho_s(\alpha) ~=~ \frac{1}{|\mathcal{P}|} \text{size}\{p \in \mathcal{P} ~|~ r_{p,s} \le \alpha\},
\]
where~$r_{p,s}$ is the performance ratio, defined as
\[
r_{p,s} ~=~ \frac{t_{p,s}}{\min\{t_{p,s} ~|~ s \in \mathcal{S}\}}.
\]
When a solver $s$ fails to satisfy the convergence test on a problem $p$, we set $\displaystyle r_{p,s} = 2 \max_{p \in \mathcal{P}, s \in \mathcal{S}} r_{p,s}$. One can plot~$\rho_s(\alpha)$ as a curve over~$\alpha$. The solver associated with the highest curve is the one with the best performance in terms of the metric chosen. In particular, the solver with the highest value of~$\rho_s(1)$ is the best in terms of efficiency, while the one with the highest value of~$\rho_s(\alpha)$, for large~$\alpha$, is the best in terms of robustness.

In this subsection, we assume that a neural network equipped with an activation function plays the role of a solver. Considering the five activation functions above (i.e., ReLU, ELU, SiLU, Sigmoid, and Tanh) leads to five different neural networks. When using the training dataset, for each test problem~$p$, the performance measure~$t_{p,s}$ is given by the number of iterations required for the training of each neural network~$s$ to achieve a point~$w$ that satisfies the following convergence test
\begin{equation}\label{eq:convergence_test}
\mathcal{L}(w^0;\mathcal{D}) - \mathcal{L}(w;\mathcal{D}) ~\ge~ (1 - \tau) (\mathcal{L}(w^0;\mathcal{D}) - \mathcal{L}_L(\mathcal{D})),
\end{equation}
where $\tau > 0$ is a convergence tolerance, $w^0$ is the initial vector of weights for each neural network, and $\mathcal{L}_L(\mathcal{D})$ is the lowest value of~$\mathcal{L}(w;\mathcal{D})$ obtained by any neural network in~$\mathcal{S}$. When using the testing dataset, $t_{p,s}$ represents the number of iterations required to satisfy the convergence test~\eqref{eq:convergence_test} with the empirical risk function evaluated over the testing dataset.

    \begin{figure}
    \centering
          \includegraphics[scale=0.45]{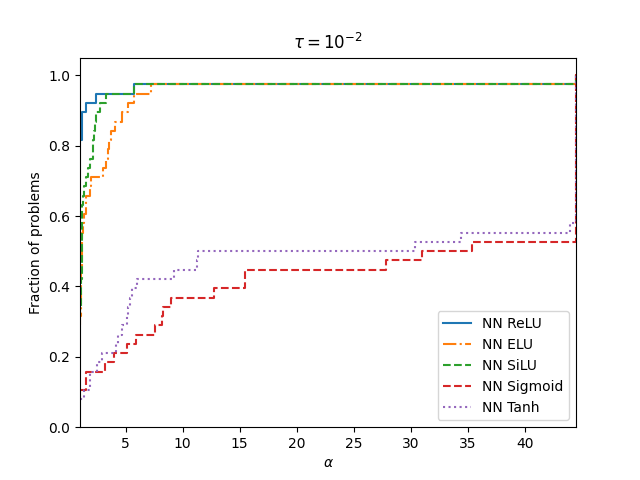} 
          \includegraphics[scale=0.45]{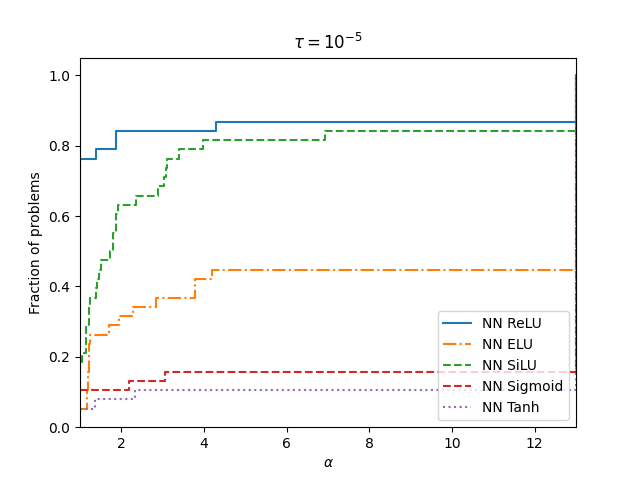} 
          \includegraphics[scale=0.45]{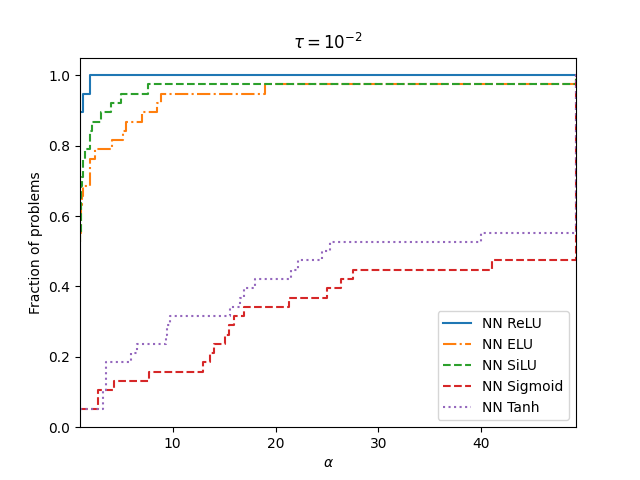} 
          \includegraphics[scale=0.45]{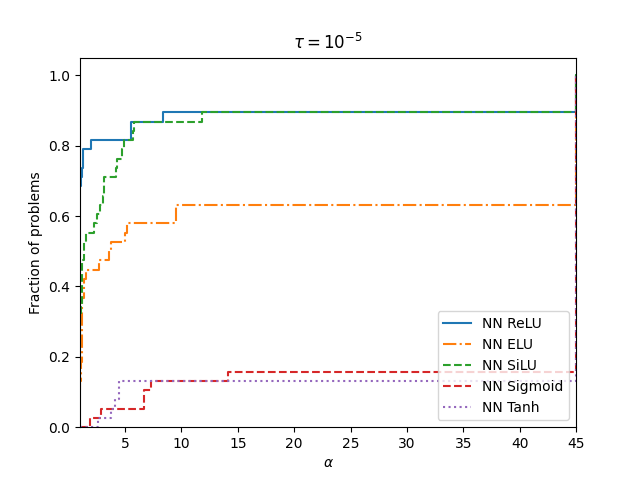}
          \includegraphics[scale=0.45]{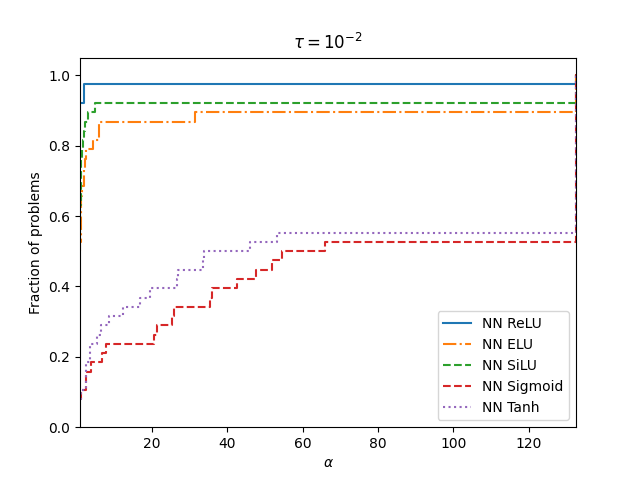} 
          \includegraphics[scale=0.45]{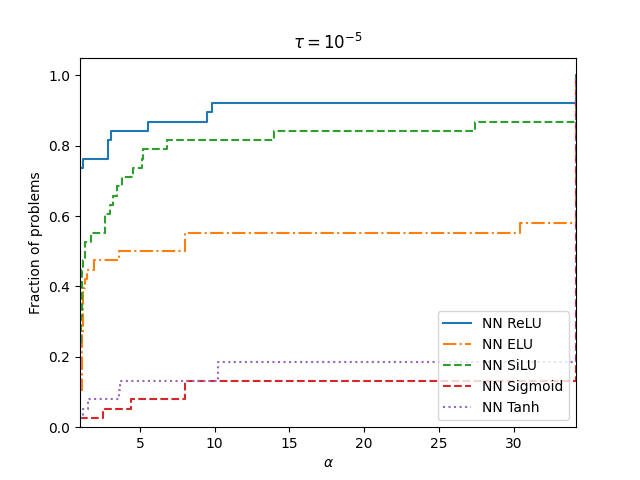}
          \caption{Performance profiles on the training dataset for $n = 20$ (first row), $n = 40$ (second row), and $n = 60$ (third row) with $\tau = 10^{-2}$ and $\tau = 10^{-5}$ for five neural networks with different activation functions on the set of~38 problems from~CUTEst. }\label{fig:activations_1}
    \end{figure}

    \begin{figure}
    \centering
          \includegraphics[scale=0.45]{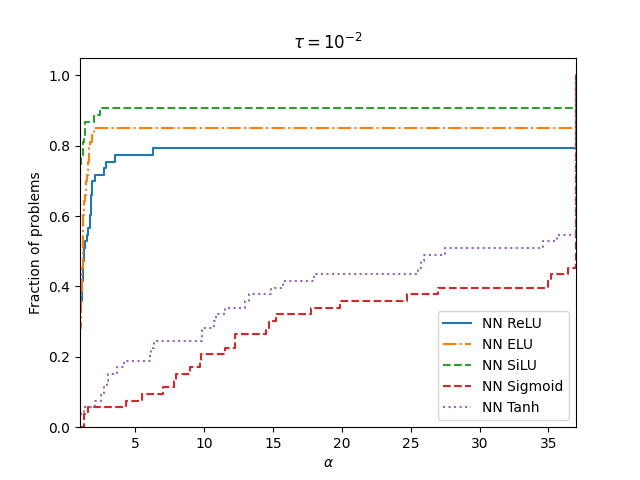} 
          \includegraphics[scale=0.45]{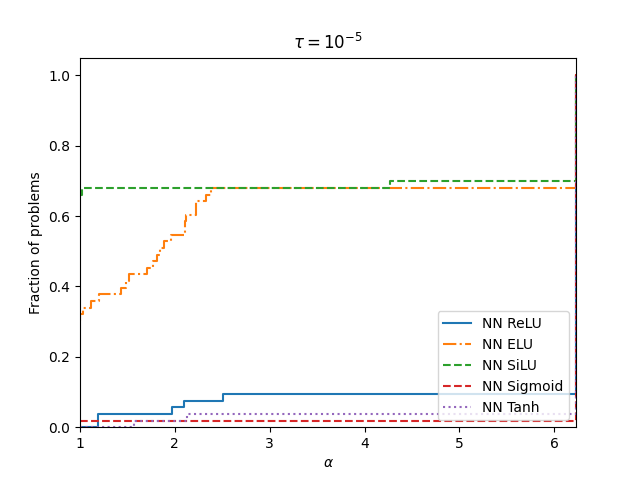}  
    \caption{Performance profiles on the training dataset with $\tau = 10^{-2}$ and $\tau = 10^{-5}$ for five neural networks with different activation functions on the set of~53 problems from CUTEst. }\label{fig:activations_2}
    \end{figure}

    \begin{figure}
    \centering
          \includegraphics[scale=0.45]{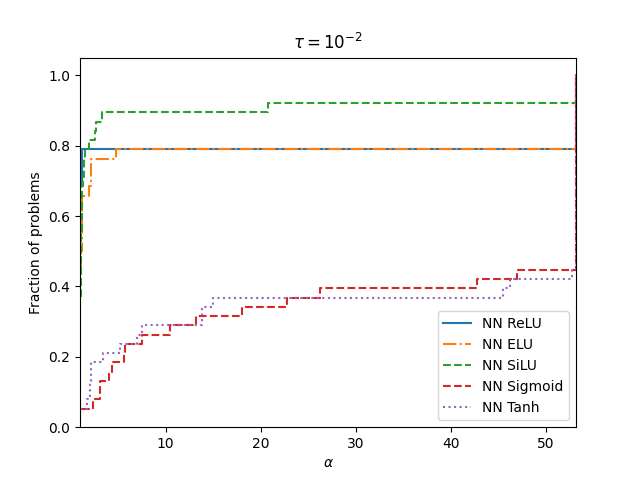} 
          \includegraphics[scale=0.45]{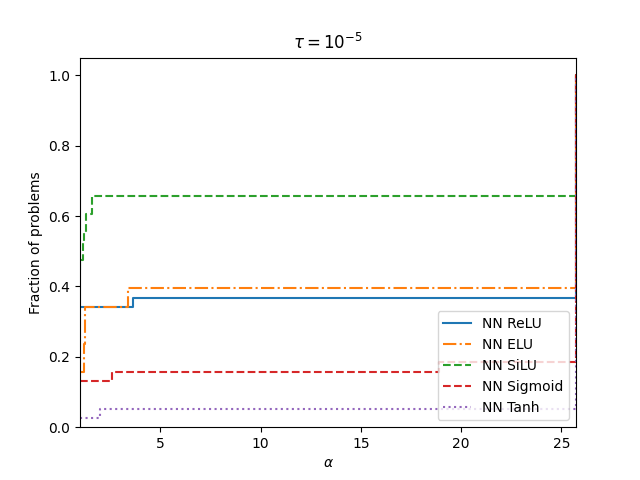} 
          \includegraphics[scale=0.45]{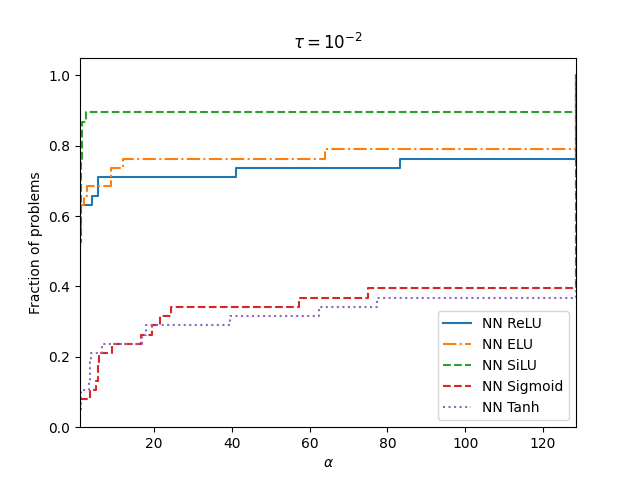} 
          \includegraphics[scale=0.45]{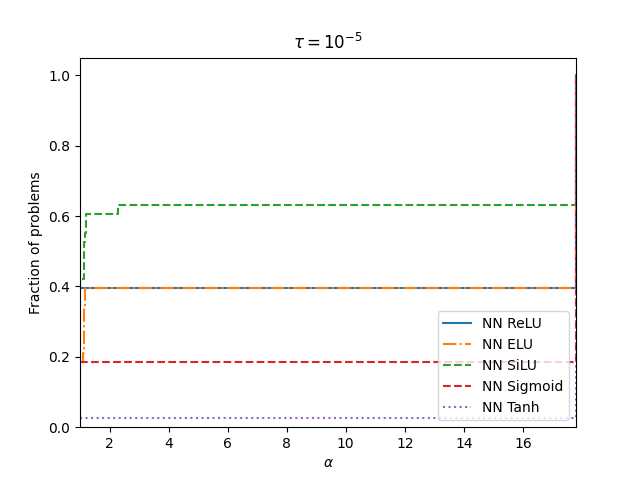}
          \includegraphics[scale=0.45]{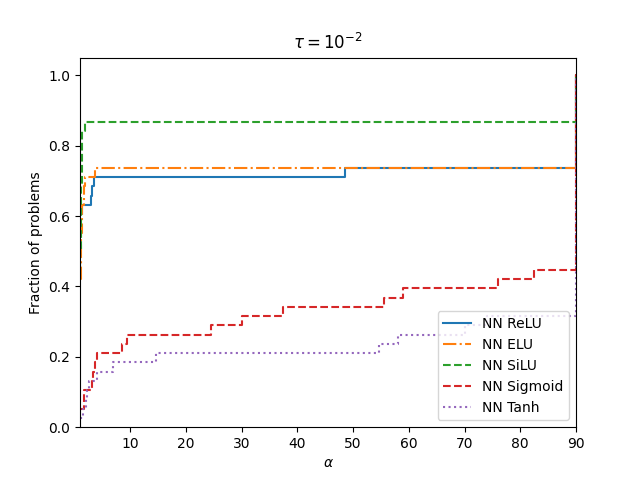} 
          \includegraphics[scale=0.45]{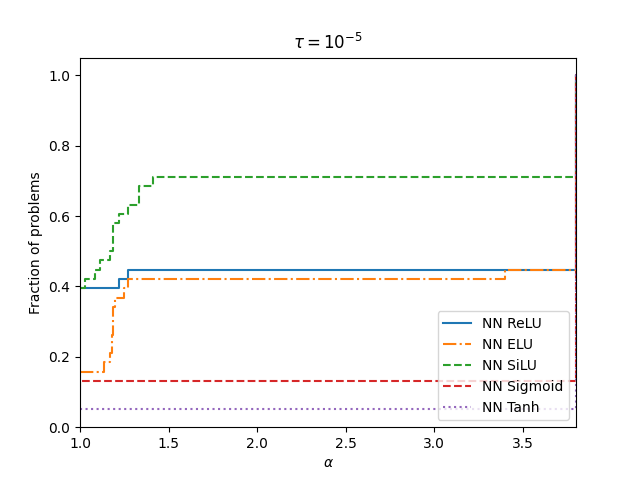}
          \caption{Performance profiles on the testing dataset for $n = 20$ (first row), $n = 40$ (second row), and $n = 60$ (third row) with $\tau = 10^{-2}$ and $\tau = 10^{-5}$ for five neural networks with different activation functions on the set of~38 problems from~CUTEst. }\label{fig:activations_test_1}
    \end{figure}

    \begin{figure}
    \centering
          \includegraphics[scale=0.45]{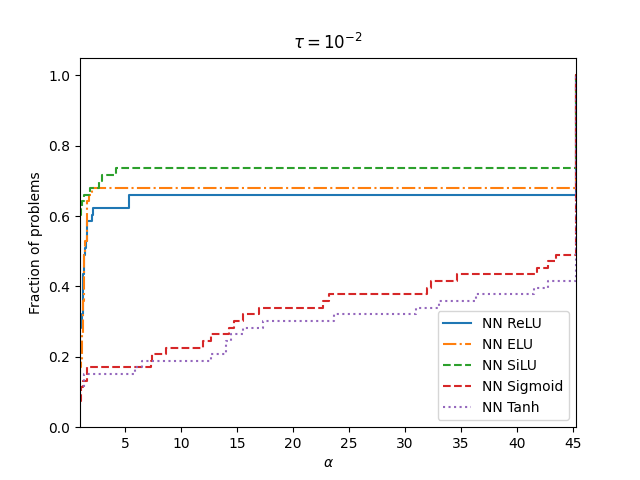} 
          \includegraphics[scale=0.45]{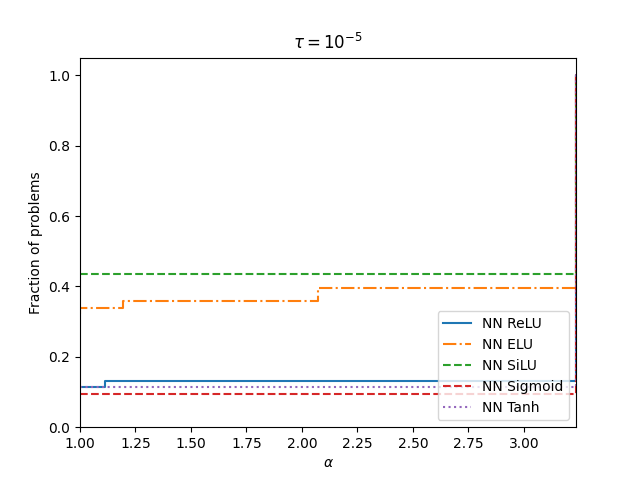}  
    \caption{Performance profiles on the testing dataset with $\tau = 10^{-2}$ and $\tau = 10^{-5}$ for five neural networks with different activation functions on the set of~53 problems from CUTEst. }\label{fig:activations_test_2}
    \end{figure}

The numerical results of our experiments were obtained using the architecture that led to the best results for all five neural networks, i.e., a feedforward neural network architecture with~2 hidden layers and~$4n$ neurons for each layer, where~$n$ is the dimension of the test problem.  We trained the five networks using Adam optimizer~\cite{DPKingma_JBa_2014} with~300 epochs.
% Such a stepsize was chosen by performing a grid search over the set $\{10^{-1},10^{-2},10^{-3},10^{-4}\}$.
We employed a dynamic learning rate adjustment strategy. Specifically, we used the ReduceLROnPlateau callback, a popular technique in deep learning~\cite{AAlKababji_etal_2022}, which led to better performance than using fixed or other decaying stepsize strategies. The learning rate was reduced by a factor of 0.8 when the empirical risk function over the testing dataset ceased to decrease, indicating a potential convergence slowdown. We configured the callback setting the mode parameter equal to~'min', as we aimed to minimize the testing empirical risk function. Additionally, a patience value of~15 epochs was set, which defines the number of epochs with no improvement on the testing empirical risk function before the learning rate reduction is triggered.
The training dataset~$\mathcal{D}$ was composed of~$N_1 = N + 1 = (n+1)(n+2)/2$ points randomly generated according to a uniform distribution in a ball~$B(x^0;1)$, where $x^0$ is the initial point provided by~CUTEst, and we refer to Section~\ref{sec:basis_functions} (where we will see that~$(n+1)(n+2)/2$ is the number of points required for determined quadratic interpolation). The testing dataset was composed of~$0.2(n+1)(n+2)/2$ points randomly generated according to a uniform distribution in the same ball~$B(x^0;1)$ used for the training dataset. The minibatch size for training was equal to~16, 32, and~64 for the dimensions~20, 40, and~60, respectively, for the first set of problems. For the second set, we selected the minibatch as the closest power of~2 to~0.05 times the size of the dataset. To ensure practicality and avoid extreme values, we constrained the mini-batch size to a minimum of~2 and a maximum of~64. %~0.05$(n+1)(n+2)/2$. 
The value of~$w^0$ was initialized using specific weight initialization techniques tailored to the activation function used. For ReLU and its variants, we applied the~He initialization \cite{he2015delving} method. For Sigmoid and Tanh, we utilized Xavier/Glorot initialization \cite{glorot2010understanding}. These initialization strategies were chosen to address the challenges associated with each activation function and promote stable and efficient training.

For numerical stability and consistency, we normalized both the training dataset~$\mathcal{D}$ and the testing dataset. This involved shifting each sample by $-x^0$ and scaling each sample by~$\Delta = \max_{1 \le i \le N} \|x^i - x^0\|$. As a result, the transformed dataset was contained within a ball of radius one centered at the origin, ensuring that at least one point lies on the boundary of the ball: 
\begin{equation}\label{eq:shifting_scaling_NN}
\left\{0,\frac{x^1-x^0}{\Delta},\ldots,\frac{x^N-x^0}{\Delta}\right\} ~\subset~ B(0;1).
\end{equation}
Note that normalizing a dataset can help mitigate the impact of the vanishing gradient issue (see~\cite{SIoffe_CSzegedy_2015}). To use the scaling and shifting in~\eqref{eq:shifting_scaling_NN}, we solved the training problem~\eqref{prob:testprob} by replacing the function~$f_{\NN}$ in~$\mathcal{L}(w;\mathcal{D})$ with  
    \begin{equation}\label{eq:NN_hat}
        \hat{f}_{\NN}(x;w) = f_{\NN}((x-x^0)/\Delta;w).
    \end{equation}
Similarly, we used~\eqref{eq:NN_hat} when evaluating the empirical risk function at points in the testing dataset.

The numerical results are reported in Figures~\ref{fig:activations_1}--\ref{fig:activations_2} for the training dataset and Figures~\ref{fig:activations_test_1}--\ref{fig:activations_test_2} for the testing dataset. The values of~$\tau$ were chosen from~$\{10^{-2}, 10^{-5}\}$. Regardless of the dimension~$n$, ReLU, ELU, and~SiLU have superior performance compared to the sigmoidal activation functions on both sets of test problems in terms of both efficiency (see the value of~$\rho_s(1)$) and robustness (see the value of~$\rho_s(\alpha)$ for large~$\alpha$). SiLU achieves the best performance on the testing dataset for both values of~$\tau$, followed by~ELU and~ReLU. On the training data, ReLU performs best on the set of~38 problems but is outperformed by both~SiLU and~ELU on the set of~53 problems. 
From these figures, we can also see that the sigmoidal activation functions are not good activation functions for approximation, which is consistent with what is commonly observed in classification tasks, where the performance of ReLU and its variants is often found to be superior~\cite{GBingham_RMiikkulainen_2020}. 

%%%%%%%%%%%%%%%%%%%%%%%%%%%%%%%%%%%%%%%%%%%%%%%%%%%%%%%%%%%%%%%%%%%%%%%%%%%%%%%%%%%%%%%%%
% \section{The use of an activation function as a basis function} 
\section{The limitation of neural nets for approximation} \label{sec:basis_functions}
%%%%%%%%%%%%%%%%%%%%%%%%%%%%%%%%%%%%%%%%%%%%%%%%%%%%%%%%%%%%%%%%%%%%%%%%%%%%%%%%%%%%%%%%%

    We start this section by reviewing the basic concepts of polynomial interpolation and regression in Subsection~\ref{subsec:interpolation_regression}. Then, we show that using the composition of an activation function with the natural basis can lead to better approximations of function values, gradients, and Hessians compared to the natural basis (see Subsection~\ref{subsec:comparison_activation}). Lastly, we assess the accuracy of the function value, gradient, and Hessian approximations obtained when using neural networks (see Subsection~\ref{subsec:comparison_NN}). 

    %%%%%%%%%%%%%%%%%%%%%%%%%%%%%%%%%%%%%%%%%%%%%%%%%%%%%%%%%%%%%%%%%%%%%%%%%%%%%%%%%%%%%%%%%
    \subsection{Review of polynomial interpolation and regression}\label{subsec:interpolation_regression}
    %%%%%%%%%%%%%%%%%%%%%%%%%%%%%%%%%%%%%%%%%%%%%%%%%%%%%%%%%%%%%%%%%%%%%%%%%%%%%%%%%%%%%%%%%

    Polynomial interpolation and regression models are frequently used in DFO methods to approximate computationally expensive objective functions in black-box problems by using polynomials~\cite{ARConn_PhLToint_1996,MJDPowell_2001,SMWild_RGRegis_CAShoemaker_2007,ARConn_KScheinberg_LNVicente_2006,ARConn_KScheinberg_LNVicente_2005,ARConn_KScheinberg_LNVicente_2009,ASBandeira_KScheinberg_LNVicente_2012}. If the number of sample points used to build an interpolation model is less than its degrees of freedom, the resulting model is underdetermined, and such a situation may arise when functions are too expensive to allow collecting a large number of function values. If there are more sample points than degrees of freedom, one can use a regression model (overdetermined interpolation), which is particularly useful when function values are noisy. If the sample size matches the number of degrees of freedom, the interpolation model is determined. The nonlinear class of models that is often used for both polynomial interpolation and regression are quadratics~\cite{DWinfield_1969,ARConn_PhLToint_1996,MJDPowell_2002,MJDPowell_2004,ASBandeira_KScheinberg_LNVicente_2012}. As opposed to linear models, quadratic models are able to capture the curvature of the function being approximated. 
    
    Let us denote as~$\mathcal{P}_n^d$ the space of polynomials of degree less than or equal to $d$ in~$\mathbb{R}^n$. Let~$q_1 = q + 1$ be the dimension of this space, and let $\phi = \{ \phi_0(x),\phi_1(x), \ldots, \phi_q(x)\}$ be a basis for~$\mathcal{P}_n^d$ (each~$\phi_j(x)$, with~$j \in \{0,\ldots,q\}$, is referred to as a basis function). 
    % We assume $\phi$ to be a polynomial basis, and we represent $\phi (x) = [\phi_0(x),\phi_1(x), \ldots, \phi_q(x)]^\top$ as a vector in~$\mathbb{R}^{q_1}$. 
    Since $\phi$ is a basis in~$\mathcal{P}_n^d$, given real scalars $\alpha_j$, any polynomial~$m(x) \in \mathcal{P}_n^d$ can be written as~$m(x) = \sum_{j=0}^{q} \alpha_j \phi_j(x)$. The most common polynomial basis is the natural basis~$\bar{\phi}$, which is the basis of polynomials in the Taylor expansion
    \begin{equation}\label{eq:natural_basis}
    \bar{\phi} ~=~ \{1,x_1,x_2,\ldots,x_n,x_1^2/2,x_1x_2,\ldots,x^{d-1}_{n-1}x_n/(d-1)!,x_n^d/d!\}.
    \end{equation}
    
    Throughout this section, we refer to $Y = \{y^0,y^1,\ldots,y^N \} \subset \mathbb{R}^n$ as a sample set, which is a set of $N_1 = N+1$ points where the objective function is evaluated. In the linear case (i.e.,~$d = 1$), the dimension of~$\mathcal{P}_n^d$ is~$q_1 = n+1$. In the quadratic case (i.e.,~$d = 2$), the dimension of~$\mathcal{P}_n^d$ is~$q_1 = (n+1)(n+2)/2$. Underdetermined quadratic interpolation models are based on sample sets of size~$N_1$ such that~$n+1 < N_1 < (n+1)(n+2)/2$. Quadratic regression models are based on sample sets of size~$N_1 > (n+1)(n+2)/2$. When $N_1 = (n+1)(n+2)/2$, one has determined quadratic interpolation. 
    
    Let us now formally describe the optimization problems arising in polynomial interpolation and regression. Denoting the function to approximate as~$f: \mathbb{R}^n \to \mathbb{R}$, let~$f(Y)$ represent the vector whose elements are~$f(y^i)$, with~$i \in \{0, \ldots, N\}$. The goal of polynomial interpolation is to determine the coefficients~$\alpha_0, \ldots, \alpha_q$ of a polynomial~$m(x)$ such that $m(y^i) = \sum_{j=0}^{q} \alpha_j \phi_j(y^i) = f(y^i)$, for all~$i \in \{0, \ldots, N\}$, see~\cite{ARConn_KScheinberg_LNVicente_2006}. To ensure that $m(x)$ interpolates the function $f$ at the points in $Y$, one can solve the following linear system, written in matrix form
    \begin{equation}\label{prob:interpolation}
    M(\phi,Y) \alpha_{\phi} ~=~ f(Y),
    \end{equation}
    where
    \begin{equation*}
        M(\phi, Y)=\left[\begin{array}{cccc}
        \phi_0\left(y^0\right) & \phi_1\left(y^0\right) & \cdots & \phi_q\left(y^0\right) \\
        \phi_0\left(y^1\right) & \phi_1\left(y^1\right) & \cdots & \phi_q\left(y^1\right) \\
        \vdots & \vdots & \vdots & \vdots \\
        \phi_0\left(y^N\right) & \phi_1\left(y^N\right) & \cdots & \phi_q\left(y^N\right)
        \end{array}\right], \ \alpha_{\phi} = \left[\begin{array}{c} \alpha_0 \\ \alpha_1 \\ \vdots \\ \alpha_q \end{array}\right], \text{ and } f(Y) = \left[\begin{array}{c} f(y^0) \\ f(y^1) \\ \vdots \\ f(y^N) \end{array}\right].
    \end{equation*}
    
    The goal of polynomial regression is to determine the coefficients $\alpha_0, \ldots, \alpha_q$ of a polynomial~$m(x)$ to solve system~\eqref{prob:interpolation} in the least-squares sense by minimizing $\| M(\phi,Y)\alpha_{\phi} - f(Y)\|^2$, see~\cite{ARConn_KScheinberg_LNVicente_2005}. When~$N = q$, interpolation and regression yield the same results. Following~\cite{ARConn_KScheinberg_LNVicente_2009}, we say that a sample set $Y$ is poised for polynomial interpolation in~$\mathbb{R}^n$ if $M(\phi,Y)$ is square~($N = q$) and non-singular. We say that a sample set~$Y$ is poised for polynomial least-squares regression in~$\mathbb{R}^n$ if~$M(\phi,Y)$ has full-column rank.
    
    For numerical reasons, it is convenient to shift~$Y$ by~$-y^0$ so that the new set~$\{0,y^1-y^0,\ldots,y^N-y^0\}$ contains the origin. In optimization methods based on polynomial interpolation or regression, the current best iterate is usually selected as such a point~$y^0$, see~\cite{ARConn_KScheinberg_LNVicente_2009}. As in~\eqref{prob:testprob}--\eqref{eq:shifting_scaling_NN}, one can then scale the resulting sample set by~$\Delta = \max_{1 \le i \le N} \|y^i - y^0\|$ so that the new set is contained in a ball of radius one centered at the origin, with at least one point on the boundary of the ball~ $B(0;1)$. 
    Throughout this paper, when using interpolation or regression, we will use such a shifting and scaling. Note that this can be achieved by introducing a polynomial~$\hat{m}(x)$ such that
    \begin{equation}\label{eq:polynomial_hat}
        \hat{m}(x) = m((x-y^0)/\Delta)
    \end{equation}
    and then performing interpolation or regression on the original sample set to ensure~$\hat{m}(y^i) = f(y^i)$, $i \in \{0,\ldots,N\}$, exactly (in the interpolation case) or in the least-squares sense (in the regression case). 

    %%%%%%%%%%%%%%%%%%%%%%%%%%%%%%%%%%%%%%%%%%%%%%%%%%%%%%%%%%%%%%%%%%%%%%%%%%%%%%%%%%%%%%%%%
    \subsection{The use of an activation function as a basis function}\label{subsec:comparison_activation}
    %%%%%%%%%%%%%%%%%%%%%%%%%%%%%%%%%%%%%%%%%%%%%%%%%%%%%%%%%%%%%%%%%%%%%%%%%%%%%%%%%%%%%%%%%
    
    In this subsection, we explore the use of bases obtained from the composition of the basis functions in the natural basis~\eqref{eq:natural_basis} with an activation function~$s: \mathbb{R} \to \mathbb{R}$. We focus on the quadratic case (i.e.,~$d = 2$ in~\eqref{eq:natural_basis}), which is widely used in practice~\cite{DWinfield_1969,ARConn_PhLToint_1996,MJDPowell_2002,MJDPowell_2004,ASBandeira_KScheinberg_LNVicente_2012}. The first type of basis\footnote{Note that although~\eqref{eq:mod_natural_basis} and~\eqref{eq:modmod_natural_basis} are motivated from the natural basis for the space of polynomials~$\mathcal{P}_n^2$, they do not form bases for such a space due to the use of an activation function. Moreover,~\eqref{eq:modmod_natural_basis} has only $3n+1$ elements.} we consider is given by
    \begin{equation}\label{eq:mod_natural_basis}
    \tilde{\phi} ~=~ \{1,x_1,x_2,\ldots,x_n,x_1^2/2,s(x_1x_2),\ldots,s(x_{n-1}x_n),x_n^2/2\},
    \end{equation}
    which is obtained from the natural basis~\eqref{eq:natural_basis} by applying~$s$ to each cross-term~$x_i x_j$, with~$\{i,j\} \subseteq \{1,\ldots,n\}$ and $i \ne j$. The second type of basis is obtained from~\eqref{eq:mod_natural_basis} by removing the cross-terms $x_i x_j$ and adding extra terms $s(x_i)$, for all~$i \in \{1, \ldots, n\}$, leading to    \begin{equation}\label{eq:modmod_natural_basis}
    \hat{\phi} ~=~ \{1,x_1,x_2,\ldots,x_n,s(x_1),s(x_2),\ldots,s(x_n),x_1^2/2,\ldots,x_n^2/2\}.
    \end{equation}   
    We point out that we also considered the bases that can be obtained from~\eqref{eq:mod_natural_basis} and~\eqref{eq:modmod_natural_basis} by applying~$s$ to each quadratic term~$x_i^2/2$, with $i \in \{1, \ldots, n\}$. However, we have not included them in this paper as they did not show any improvement in performance compared to the natural basis. 
    
    In order to have a benchmark to compare bases~\eqref{eq:mod_natural_basis} and~\eqref{eq:modmod_natural_basis} with, we considered a third type of basis consisting of radial basis functions~\cite{MDBuhmann_2003,DBMcDonald_WJGrantham_2007,HMGutmann_2001}. Such a basis can be written as follows
    \begin{alignat}{2}\label{eq:RBF}
    \doublehat{\phi} 
    % &~=~ \{\phi_0(x), \phi_1(x), \ldots, \phi_N(x)\} \\ 
    &~=~ \{h(\Vert x - y^0\Vert), \, h(\Vert x - y^1\Vert), \ldots, h(\Vert x - y^N\Vert), 1, x_1, \ldots, x_n\},
    \end{alignat}
    where $h: \mathbb{R} \to \mathbb{R}$ and we are using constant and linear terms as suggested in~\cite{ARConn_KScheinberg_LNVicente_2009}. Some of the most popular radial basis functions are cubic~$h(z) = z^3$, Gaussian~$h(z) = e^{-(z^2/\rho^2)}$, multiquadric~$h(z) = (z^2 + \rho^2)^{(3/2)}$, and inverse multiquadric~$h(z) = (z^2 + \rho^2)^{-(1/2)}$, where~$\rho^2$ is a positive constant.
    When using basis~\eqref{eq:RBF}, given vectors of real scalars denoted as~$\lambda = (\lambda_0, \lambda_1, \ldots, \lambda_N)$ and~$\gamma = (\gamma_{0}, \gamma_{1}, \ldots, \gamma_{n})$, the resulting model becomes~$m(x) = \sum_{j=0}^{N}\lambda_j h(\Vert x - y^j \Vert)+ \sum_{j=0}^{n}\gamma_j p_j(x)$, where each~$p_j(x)$ represents one of the constant and linear terms in~\eqref{eq:RBF}.
    To determine the coefficients of such a model, it is common to solve the solve the following linear system~\cite{ARConn_KScheinberg_LNVicente_2009,JLarson_MMenickelly_SMWild_2019}
    \begin{equation}\label{eq:RBFsystem}
        \left[\begin{array}{cc}
        \Phi & P \\
        P^\top & 0 
        \end{array}\right] \left[\begin{array}{c} 
        \lambda \\ \gamma 
        \end{array}\right] 
        = 
        \left[\begin{array}{c} 
        f(Y) \\ 0
        \end{array}\right],
    \end{equation}
    where~$\Phi_{ij} = h(\Vert y^i - y^j\Vert)$, for all~$i,j \in \{0,1,\ldots,N\}$, and~$P_{ij} = p_j(y^i)$, for all~$i \in \{0,1,\ldots,N\}$ and~$j \in \{0,\ldots,n\}$. The system~\eqref{eq:RBFsystem} ensures that the interpolation conditions~$m(y^i) = f(y^i)$ are satisfied for all~$i \in \{0,\ldots,N\}$ and enforces~$\sum_{i=0}^{N} \lambda_i p_j(y^i) = 0$, for all~$j \in \{0,\ldots,n\}$. In accordance with our notation in~\eqref{eq:polynomial_hat} for interpolation and regression, we will refer to a scaled and shifted model using radial basis functions as~$\hat{m}(x)$.
    
    Note that the number of degrees of freedom in~$m(x)$ when using~basis~\eqref{eq:modmod_natural_basis} is $3n + 1$, which is less than the number required for determined quadratic interpolation (i.e.,~$(n+1)(n+2)/2$).
    The same remark applies to~$\hat{m}(x)$, defined in~\eqref{eq:polynomial_hat}. Since we want to test bases~\eqref{eq:mod_natural_basis} and~\eqref{eq:modmod_natural_basis} using a sample set of dimension~$(n+1)(n+2)/2$ for both, we need to solve a determined interpolation problem when using~\eqref{eq:mod_natural_basis} and a regression problem when using~\eqref{eq:modmod_natural_basis}.  Therefore, we will refer to~$\hat{m}(x)$ as an interpolation model when using~\eqref{eq:mod_natural_basis} and as a regression model when using~\eqref{eq:modmod_natural_basis}. When using basis~\eqref{eq:RBF}, we will refer to the resulting model~$\hat{m}(x)$ as an interpolation model.

    To assess the performance of bases~\eqref{eq:mod_natural_basis} and~\eqref{eq:modmod_natural_basis} for interpolation and regression, respectively, and basis~\eqref{eq:RBF} for interpolation, we used the same sets of test problems considered in Section~\ref{sec:activation_functions} and listed in Tables~\ref{Tab:40probs} and~\ref{Tab:negcurv_probs} of Appendix~\ref{app:tables}. Here, we are interested in assessing the ability of the proposed basis functions to locally approximate the function~$f$ by using an interpolation/regression model~$\hat{m}(x)$ as a surrogate model of~$f$. To perform this assessment, we compared the function value, gradient, and Hessian approximations obtained using bases~\eqref{eq:mod_natural_basis} and~\eqref{eq:modmod_natural_basis} for different choices of the activation~$s$ and basis~\eqref{eq:RBF} equipped with Gaussian radial basis functions. Similar to the experiments for Section~\ref{sec:activation_functions}, we built the sample set~$Y$ by randomly generating~$N_1 = (n+1)(n+2)/2$ points according to a uniform distribution in a ball~$B(x^0;1)$, where $x^0$ is the initial point provided by~CUTEst.
    
\begin{figure}
    \centering
    \includegraphics[scale=0.60]{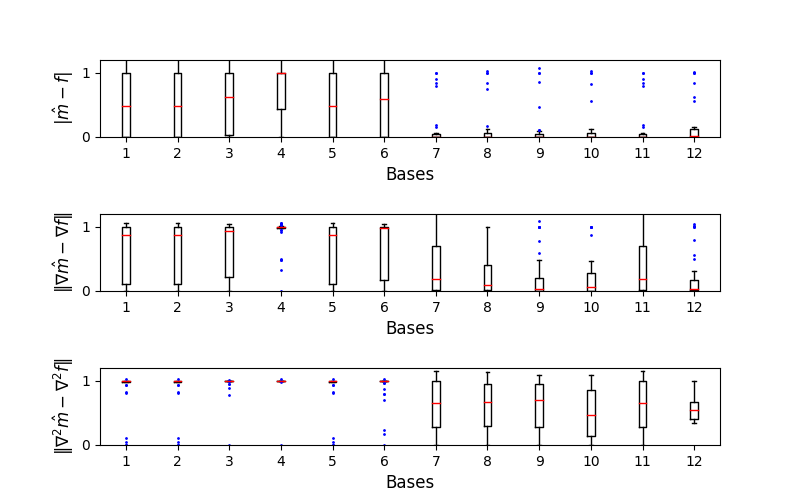}  
        \caption{Comparison of~$|\hat{m} -f|$, $\|\nabla \hat{m} -\nabla f\|$, and~$\|\nabla^2 \hat{m} -\nabla^2 f\|$ among different bases for the set of~38 problems listed in Table~\ref{Tab:40probs} of Appendix~\ref{app:tables}.}\label{fig:func_38probs}
    \end{figure}

\begin{figure}
    \centering
    \includegraphics[scale=0.60]{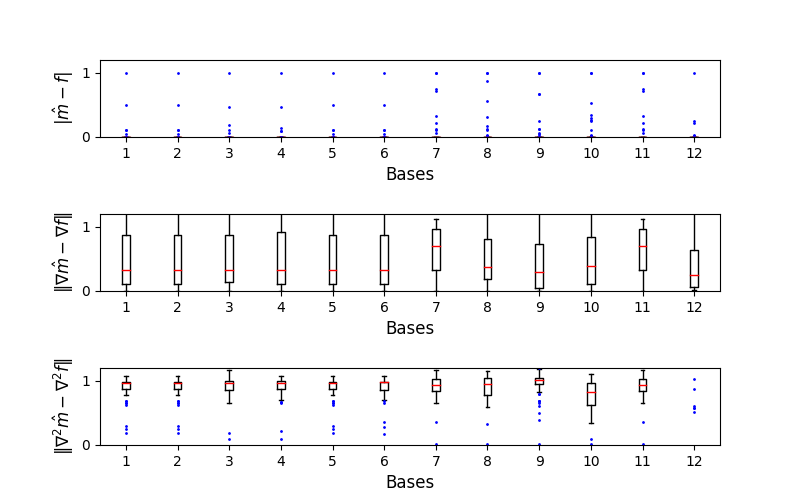}  
        \caption{Comparison of~$|\hat{m} -f|$, $\|\nabla \hat{m} -\nabla f\|$,~$\|\nabla^2 \hat{m} -\nabla^2 f\|$ among different bases for the set of~53 problems listed in Table~\ref{Tab:negcurv_probs} of Appendix~\ref{app:tables}.}\label{fig:func_negcurvprobs}
    \end{figure}

    The results of our numerical experiments are illustrated in Figures~\ref{fig:func_38probs} and~\ref{fig:func_negcurvprobs}, which report box plots that allow us to compare the natural basis~\eqref{eq:natural_basis}, bases~\eqref{eq:mod_natural_basis} and~\eqref{eq:modmod_natural_basis} when the activation function~$s$ is either~ReLU, ELU, SiLU, Sigmoid, or Tanh, and basis~\eqref{eq:RBF}. In particular, Figure~\ref{fig:func_38probs} corresponds to the set of~38 problems (with~$n = 20$), while Figure~\ref{fig:func_negcurvprobs} pertains to the set of~53 problems. Recalling the definition of~$\hat{m}(x)$ in~\eqref{eq:polynomial_hat}, the y-axis of the upper plot in Figures~\ref{fig:func_38probs} and~\ref{fig:func_negcurvprobs} represents the value of the following metric
    \begin{equation}\label{eq:metric_1}
    |\hat{m} - f| \; := \; \frac{|\hat{m}(x^0) - f(x^0)|}{\max\{|\hat{m}(x^0)|,|f(x^0)|\}},
    \end{equation}
    where the max function is used in the denominator to avoid a division by zero when either~$\hat{m}(x^0)$ or $f(x^0)$ is equal to zero (in the experiments, such terms are never both equal to zero). Such a metric evaluates the function value approximation error by comparing the value of~$f$ at the initial point provided by CUTEst with the value of~$\hat{m}$ at the same point.
    We adopted similar metrics to evaluate the gradient and Hessian approximation errors. Specifically, the y-axes of the middle and lower plots in Figures~\ref{fig:func_38probs} and~\ref{fig:func_negcurvprobs} represent the values of
    \begin{equation}\label{eq:metric_2}
    \Vert \nabla \hat{m} - \nabla f \Vert \; := \; \frac{\|\nabla \hat{m}(x^0) - \nabla f(x^0)\|}{\max\{\|\nabla \hat{m}(x^0)\|,\|\nabla f(x^0)\|\}} 
    \end{equation}
    and
    \begin{equation}\label{eq:metric_3}
    \Vert \nabla^2 \hat{m} - \nabla^2 f \Vert \; := \; \frac{\Vert\nabla^2 \hat{m}(x^0) - \nabla^2 f(x^0)\|}{\max\{\|\nabla^2 \hat{m}(x^0)\|,\|\nabla^2 f(x^0)\|\}},  
    \end{equation}
    respectively. In all the plots, the x-axis corresponds to the natural basis~\eqref{eq:natural_basis}, the bases derived from~\eqref{eq:mod_natural_basis} and~\eqref{eq:modmod_natural_basis}, and basis~\eqref{eq:RBF} according to the following notation:
    \begin{center}
    \begin{tabular}{cl}
        1: & \eqref{eq:natural_basis}, \\
        2: & \eqref{eq:mod_natural_basis} with ReLU, \\
        3: & \eqref{eq:mod_natural_basis} with ELU, \\
        4: & \eqref{eq:mod_natural_basis} with SiLU, \\
        5: & \eqref{eq:mod_natural_basis} with Sigmoid, \\
        6: & \eqref{eq:mod_natural_basis} with Tanh, \\
        7: & \eqref{eq:modmod_natural_basis} with ReLU, \\
        8: & \eqref{eq:modmod_natural_basis} with ELU, \\
        9: & \eqref{eq:modmod_natural_basis} with SiLU,\\
        10: & \eqref{eq:modmod_natural_basis} with Sigmoid, \\
        11: & \eqref{eq:modmod_natural_basis} with Tanh,\\
        12: & \eqref{eq:RBF}.
    \end{tabular}
    \end{center} 
    Each box plot illustrates the values of a metric from~\eqref{eq:metric_1}--\eqref{eq:metric_3} obtained over all the problems considered in the corresponding figure. We recall that in a box plot, the horizontal line within the rectangle is the median of the set of values, and the upper and lower lines are the medians of the upper and lower halves of the same set of values, respectively. The circles represent outliers. To compute the boxplots, we restrict the values of metrics~\eqref{eq:metric_1}--\eqref{eq:metric_3} to the range~$[0,5]$ in order to exclude extreme outliers. In the figures, we limit the y-axis to the range~$[0,1]$.

    Figure~\ref{fig:func_38probs} shows that the bases derived from~\eqref{eq:modmod_natural_basis} allow one to obtain the most accurate model~$\hat{m}$ in terms of function value, gradient, and Hessian approximations on the set of~38 problems. Basis~\eqref{eq:RBF} exhibits a similar performance to~\eqref{eq:modmod_natural_basis}, although it is slightly worse in terms of Hessian approximation. Note that the values of metrics~\eqref{eq:metric_1}--\eqref{eq:metric_3} obtained using the bases derived from~\eqref{eq:mod_natural_basis} are comparable to those achieved using the natural basis~\eqref{eq:natural_basis} except when using SiLU as an activation function, and we can observe that both~\eqref{eq:natural_basis} and~\eqref{eq:mod_natural_basis} lead to quite inaccurate function value, gradient, and Hessian approximations (this is somehow expected given that the radius of the ball is equal to~1, and we observed that the inaccuracy reduces if we take a smaller ball). 
    
    When considering the set of~53 problems, Figure~\ref{fig:func_negcurvprobs} shows that~\eqref{eq:natural_basis} and~\eqref{eq:mod_natural_basis} have similar performance to~\eqref{eq:modmod_natural_basis} in terms of function value, gradient, and Hessian approximations. Therefore, such results suggest that the cross-terms~$x_i x_j$ in~\eqref{eq:natural_basis} and~\eqref{eq:mod_natural_basis} might not be necessary, regardless of the presence of negative curvature in the functions being approximated.
    We recall that employing a sample set of size~$N_1$ to assess the performance of both~\eqref{eq:mod_natural_basis} and~\eqref{eq:modmod_natural_basis} required us to solve a fully determined interpolation problem when using~\eqref{eq:mod_natural_basis} and a regression problem when using~\eqref{eq:modmod_natural_basis}. Although basis~\eqref{eq:RBF} performs well in terms of function value and gradient approximations, it is not able to capture curvature information with the same level of accuracy. We point out that we repeated the analysis by replacing~$x^0$ with points located between the origin and $x^0$ itself, and this did not change our conclusions. 

    Our conclusion is that the use of activation functions in quadratic interpolation and regression seems to waive the necessity of including cross terms.
    
    %%%%%%%%%%%%%%%%%%%%%%%%%%%%%%%%%%%%%%%%%%%%%%%%%%%%%%%%%%%%%%%%%%%%%%%%%%%%%%%%%%%%%%%%%
    \subsection{The use of neural networks for approximation}\label{subsec:comparison_NN}
    %%%%%%%%%%%%%%%%%%%%%%%%%%%%%%%%%%%%%%%%%%%%%%%%%%%%%%%%%%%%%%%%%%%%%%%%%%%%%%%%%%%%%%%%%

    In this subsection, we assess the accuracy of the function value, gradient, and Hessian approximations obtained when using a neural network as a surrogate model of the function~$f$. In particular, we considered five neural networks, each equipped with either ReLU, ELU, SiLU, Sigmoid, or Tanh. To conduct this analysis and ensure a fair comparison with the numerical results obtained for the interpolation/regression model in Subsection~\ref{subsec:comparison_activation}, we constructed the training dataset~$\mathcal{D} = \{(x^i,f(x^i)) ~|~ i \in \{0,\ldots,N\}\}$ using the same approach that we employed to build the sample set~$Y$. Specifically, we randomly generated~$N_1 = N + 1 = (n+1)(n+2)/2$ points according to a uniform distribution in a ball~$B(x^0;1)$, where $x^0$ is once again the initial point provided by~CUTEst.
    Then, after applying the shifting and scaling described in~\eqref{eq:shifting_scaling_NN}, we solved problem~\eqref{prob:testprob} for each neural network. For the details of the neural network architecture and the training process, we refer to Subsection~\ref{subsec:activation_functions_numerical_exp}. 
    % To ensure unbiased evaluation among the neural networks, we use the best models in terms of testing empirical risk function for the comparison.
    
\begin{figure}
    \centering
    \includegraphics[scale=0.60]{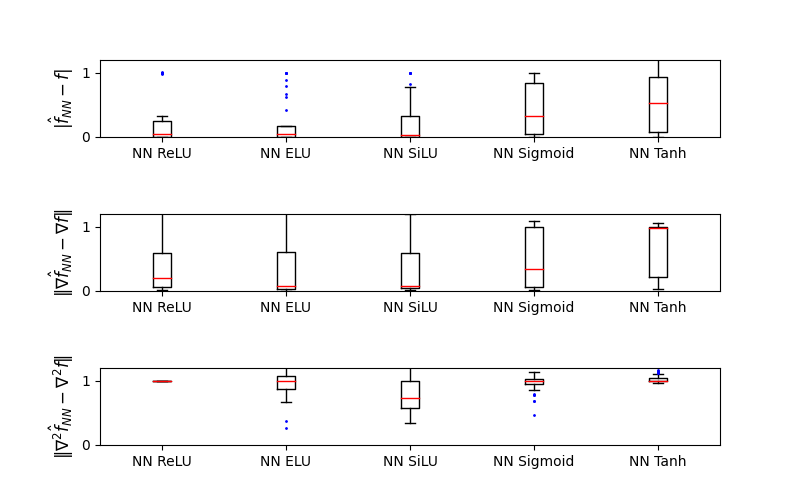}  
        \caption{Comparison of~$|\hat{f}_{\NN} -f|$, $\|\nabla\hat{f}_{\NN} -\nabla f\|$, and~$\|\nabla^2\hat{f}_{\NN} -\nabla^2 f\|$ among different activation functions for the set of~38 problems listed in Table~\ref{Tab:40probs} of Appendix~\ref{app:tables}.}\label{fig:func_38probs-NN}
    \end{figure}

\begin{figure}
    \centering
    \includegraphics[scale=0.60]{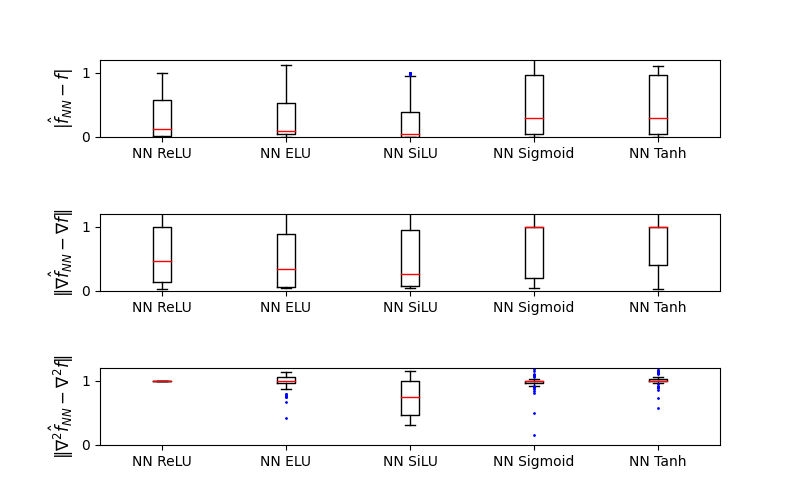}  
        \caption{Comparison of~$|\hat{f}_{\NN} -f|$, $\|\nabla\hat{f}_{\NN} -\nabla f\|$, and~$\|\nabla^2\hat{f}_{\NN} -\nabla^2 f\|$ among different activation functions for the set of~53 problems listed in Table~\ref{Tab:negcurv_probs} in Appendix~\ref{app:tables}.}\label{fig:func_negcurvprobs-NN}
    \end{figure}

    Figure~\ref{fig:func_38probs-NN} corresponds to the set of~38 problems (with~$n = 20$), while Figure~\ref{fig:func_negcurvprobs-NN} pertains to the set of~53 problems. To compare the function values, gradients, and Hessians of a neural network surrogate with those of the function~$f$, we use similar metrics to~\eqref{eq:metric_1}--\eqref{eq:metric_3}. In particular, recalling the definition of the neural network surrogate~$\hat{f}_{\NN}$ in~\eqref{eq:NN_hat}, we consider
    \begin{alignat}{2}
    |\hat{f}_{\NN} - f| \; &:= \; \frac{|\hat{f}_{\NN}(x^0) - f(x^0)|}{\max\{|\hat{f}_{\NN}(x^0)|,|f(x^0)|\}}, \label{eq:metric_1-NN} \\
    \Vert \nabla \hat{f}_{\NN} - \nabla f \Vert \; &:= \; \frac{\|\nabla \hat{f}_{\NN}(x^0) - \nabla f(x^0)\|}{\max\{\|\nabla \hat{f}_{\NN}(x^0)\|,\|\nabla f(x^0)\|\}}, \label{eq:metric_2-NN}\\ 
    \Vert \nabla^2 \hat{f}_{\NN} - \nabla^2 f \Vert \; &:= \; \frac{\Vert\nabla^2 \hat{f}_{\NN}(x^0) - \nabla^2 f(x^0)\|}{\max\{\|\nabla^2 \hat{f}_{\NN}(x^0)\|,\|\nabla^2 f(x^0)\|\}}.\label{eq:metric_3-NN}
    \end{alignat}

    Figure~\ref{fig:func_38probs-NN} reports box plots that illustrate the comparison of metrics~\eqref{eq:metric_1-NN}--\eqref{eq:metric_3-NN} among different activation functions for the set of~38 problems. In particular, the upper and middle plots in Figure~\ref{fig:func_38probs-NN} show that~ReLU and its variants are the activation functions that yield the highest accuracy in terms of function value and gradient approximations, while the sigmoidal functions exhibit lower accuracy. Note that the value of metric~\eqref{eq:metric_2-NN} is significantly large when using the sigmoidal activations. Indeed, for most of the problems, the approximate gradients obtained using the sigmoidal activations were observed to be close to the null vector due to the vanishing gradient problem, despite the use of the shifting and scaling in~\eqref{eq:shifting_scaling_NN}. The lower plot in Figure~\ref{fig:func_38probs-NN} shows that the considered neural networks were not able to accurately approximate the curvature of any objective function regardless of the activation function used, with the exception of SiLU, which shows the best performance. The limitation in capturing curvature information was somehow expected for ReLU, which is a piecewise-linear function with no curvature, but it is less obvious for the sigmoidal activations. The upper and middle plots in Figure~\ref{fig:func_negcurvprobs-NN} confirm that ReLU and its variants are the best activation functions to approximate the function values and gradients on the set of~53 problems as well. The lower plot in Figure~\ref{fig:func_negcurvprobs-NN} demonstrates that although none of the activation functions provides very accurate Hessian approximations, SiLU exhibits a superior ability to capture curvature information. 

%%%%%%%%%%%%%%%%%%%%%%%%%%%%%%%%%%%%%%%%%%%%%%%%%%%%%%%%%%%%%%%%%%%%%%%%%%%%%%%%%%%%%%%%%
\section{The limitation of neural nets for optimizing without derivatives} \label{sec:optimizing_without_derivatives}
%%%%%%%%%%%%%%%%%%%%%%%%%%%%%%%%%%%%%%%%%%%%%%%%%%%%%%%%%%%%%%%%%%%%%%%%%%%%%%%%%%%%%%%%%

The aim of this section is to enhance the performance of a state-of-the-art derivative-free optimization~(DFO) algorithm when the gradient of the objective functions of the test problems in Tables~\ref{Tab:40probs} and~\ref{Tab:negcurv_probs} of Appendix~\ref{app:tables} is approximated using the surrogate models considered in Sections~\ref{sec:activation_functions} and~\ref{sec:basis_functions}, including neural networks. DFO is a well-studied area in the field of mathematical optimization that deals with solving problems where each evaluation of the objective function is time-consuming or computationally expensive ~\cite{ARConn_KScheinberg_LNVicente_2009,CAudet_WHare_2017,JLarson_MMenickelly_SMWild_2019,ALCustodio_KScheinberg_LNVicente_2017}. Recent research in~DFO has shown that combining a Broyden-Fletcher-Goldfarb-Shanno (BFGS) quasi-Newton update step with a finite-difference~(FD) gradient allows obtaining strong performance in the minimization of a smooth and noiseless objective function by taking advantage of the curvature of such a function~\cite{ASBerahas_etal_2019,Berahas_Cao_etal_2022,HJMShi_MQXuan_etal_2023}. 

In Subsection~\ref{subsec:DFO_surrogate}, we will present the~BFGS update step with~FD gradient, as well as its surrogate-based version, as part of the general class of methods called \texttt{Full-Low Evaluation} (FLE), which will be briefly mentioned below and reviewed in Appendix~\ref{app:FLE}. FLE methods have been proposed in~\cite{ASBerahas_OSohab_LNVicente_2023} and can be used to minimize a function~$f$ without using its derivatives, which makes this class of algorithms suitable to solve problem~\eqref{prob:DFO}. In Subsection~\ref{subsec:DFO_numerical_exp}, we will report the results of our numerical experiments, where we used surrogate models to enhance the performance of a practical~FLE method.

%%%%%%%%%%%%%%%%%%%%%%%%%%%%%%%%%%%%%%%%%%%%%%%%%%%%%%%%%%%%%%%%%%%%%%%%%%%%%%%%%%%%%%%%%
\subsection{The use of a surrogate model for optimizing without derivatives}\label{subsec:DFO_surrogate}
%%%%%%%%%%%%%%%%%%%%%%%%%%%%%%%%%%%%%%%%%%%%%%%%%%%%%%%%%%%%%%%%%%%%%%%%%%%%%%%%%%%%%%%%%

We will start this subsection by describing the $k$-th iteration of the~BFGS update step with~FD gradient in Algorithm~\ref{alg:BFGS_FD}, and then we will present its surrogate-based version in Algorithm~\ref{alg:FLE-S} below. The direction explored in Algorithm~\ref{alg:BFGS_FD} is given by~$p_k = - H_k g_k$, where~$g_k$ is the approximation of the gradient~$\nabla f(x_k)$ obtained using a forward~FD scheme at the~$k$-th iterate~$x_k$ and~$H_k$ is the~BFGS approximation of the Hessian inverse~$\nabla^2 f(x_k)$. In particular, the $i$-th component of~$g_k$ is given by
\begin{equation}\label{eq:FD}
[g_k]_i ~=~  \frac{f(x_k + h_k e_i) - f(x_k)}{h_k}, \quad \forall i \in \{1, \ldots, n\},
\end{equation}
where~$h_k$ is the~FD parameter and~$e_i \in \mathbb{R}^n$ is the $i$-th canonical vector (see Step~1 for the use of~\eqref{eq:FD} in Algorithm~\ref{alg:BFGS_FD}). To compute~$H_k$, one needs first to determine~$s_k = x_k - x_{k-1}$ and~$y_k = g_k - g_{k-1}$ (see Step~2). Then, if $s_k^\top y_k \ge \varepsilon \|s_k\| \|y_k\|$, with~$\varepsilon > 0$, one can set
\begin{equation}\label{eq:H_matrix}
H_k ~=~ \left(I-\frac{s_k y_k^{\top}}{y_k^{\top} s_k}\right) H_{k-1}\left(I-\frac{y_k s_k^{\top}}{y_k^{\top} s_k}\right)+\frac{s_k s_k^{\top}}{y_k^{\top} s_k},
\end{equation}
and $H_k = H_{k-1}$ otherwise (see Step~3). The condition on the scalar product~$s_k^\top y_k$ is necessary to maintain the positive definiteness of~$H_k$. At the first iteration, one possible choice is $H_0 = (y_0^\top s_0)/(y_0^\top y_0) I$, which ensures similarity in size between~$H_0$ and~$\nabla^2 f(x_0)^{-1}$, see~\cite{JNocedal_SJWright_2006}. 

Once the direction~$p_k$ has been computed at Step~4, one can obtain the next iterate by performing a line search at Steps~5--6 and setting~$x_{k+1} = x_{k} + \beta_k p_k$ at Step~7, where~$\beta_k$ is the value of~$\beta > 0$ that satisfies the following sufficient decrease condition by backtracking from a positive initial stepsize~$\bar{\beta}$
\begin{equation}\label{eq:FE_suff_decr_noapp}
    f(x_k + \beta p_k) ~\le~ f(x_k) + c \beta g_k^\top p_k, \quad \text{ with } c \in (0,1).
\end{equation}
The sufficient decrease condition~\eqref{eq:FE_suff_decr_noapp}, which is typical in nonlinear optimization, allows one to obtain stepsizes that are neither too large nor too small~\cite{JNocedal_SJWright_2006}. 

\begin{algorithm}[H]
\caption{FLE (only showing the FD-BFGS step)}\label{alg:BFGS_FD}
\begin{algorithmic}[1]
    %\scriptsize
    \medskip
    \item[] {\bf Input:} Iterates $x_{k-1}$ and $x_{k}$, $\varepsilon > 0$, backtracking parameters $\bar{\beta} > 0$ and $\tau \in (0,1)$. 
  \smallskip 
  \item Compute the~FD gradient~$g_k$ by using~\eqref{eq:FD}.
  \smallskip
  \item Set~$s_k = x_k - x_{k-1}$ and~$y_k = g_k - g_{k-1}$.
  \smallskip
  \item \textbf{If} $s_k^\top y_k \ge \varepsilon \|s_k\| \|y_k\|$, set~$H_k$ according to~\eqref{eq:H_matrix}, \textbf{else} $H_k = H_{k-1}$.
  \smallskip
\item Compute the direction $p_k = - H_k g_k$ and set $\beta = \bar{\beta}$ for the backtracking line-search.
\smallskip
\item {\bf While} \eqref{eq:FE_suff_decr_noapp} is false {\bf do}
\item \quad\quad Set $\beta = \tau \beta$.
\item Set $\beta_k = \beta$ and $x_{k+1} = x_k + \beta_k p_k$.
\medskip
    \item[] {\bf Output:} Iterate $x_{k+1}$.
    \par\bigskip\noindent
    \end{algorithmic}
\end{algorithm}

 Note that computing~\eqref{eq:FD} requires~$n$ additional function evaluations per iteration. Rather than incurring such a cost, we propose to build a surrogate model of the objective function~$f$ at each iteration~$k$ and use its gradient as the approximate gradient~$g_k$ for the~BFGS update step. Such a model can be fitted to points in the current dataset~$\mathcal{D}_k = \{(x_i,f(x_i)) ~|~ i < k\}$, consisting of pairs~$(x_i,f(x_i))$ associated with function evaluations done at the past iterations. As the algorithm progresses and more points are collected, the accuracy of the model in approximating the objective function improves over the iterations. Since the cost of the~FD approximation in~\eqref{eq:FD} depends on~$n$, the benefit of using the gradient of the model instead of the~FD gradient is higher on large-scale optimization problems. 
 
 As a surrogate model, we consider either an interpolation or regression model, such as the ones obtained through bases~\eqref{eq:mod_natural_basis}, \eqref{eq:modmod_natural_basis}, and~\eqref{eq:RBF} proposed in Subsection~\ref{subsec:comparison_activation}, or a neural network, like the ones used in Subsection~\ref{subsec:comparison_NN}. When employing an interpolation or regression model, the sample set used at iteration~$k$, denoted by~$Y_k$, can be obtained as a subset of the points~$x_i$ in the current dataset~$\mathcal{D}_k$, i.e., $Y_k \subseteq \{x_i ~|~ (x_i,f(x_i)) \in \mathcal{D}_k\}$. When using a neural network, one can consider the entire~$\mathcal{D}_k$ as a training set. The limitation of using an interpolation or regression surrogate in the FD-BFGS step in Algorithm~\ref{alg:BFGS_FD} is that the points generated during the backtracking line search in steps~4--7 are aligned along the direction~$p_k$ and, therefore, the resulting sample set~$Y_{k}$ might not be well poised. The limitation of using a neural network surrogate is that the points generated during the backtracking line search might not be sufficient to get an accurate approximation of the objective function, and more points need to be collected. To overcome the above limitations, one can take advantage of a direct-search step to collect additional points that are more evenly distributed throughout the space.

% %%%%%%%%%%%%%%%%%%%%%%%%%%%%%%%%%%%%%%%%%%%%%%%%%%%%%%%%%%%%%%%%%%%%%%%%%%%%%%%%%%%%%%%%%
% \subsection{Numerical experiments}\label{subsec:optimizing_without_derivatives}
% %%%%%%%%%%%%%%%%%%%%%%%%%%%%%%%%%%%%%%%%%%%%%%%%%%%%%%%%%%%%%%%%%%%%%%%%%%%%%%%%%%%%%%%%%

In fact, in our experiments, we consider the class of~FLE methods, which combines the FD-BFGS step in Algorithm~\ref{alg:BFGS_FD} with a direct-search step~\cite{ASBerahas_OSohab_LNVicente_2023}. FLE methods exhibit state-of-the-art performance by leveraging two types of iterations: on the one hand, \texttt{Full-Eval}~(FE) iterations are effective in the smooth, non-noisy case but require a large number of function evaluations (this is Algorithm~\ref{alg:BFGS_FD}); on the other hand, \texttt{Low-Eval}~(LE) iterations are cheaper in terms of function evaluations and are effective in the noisy and/or non-smooth case. The details of the~LE step (direct search) are unnecessary for our discussion as well as the switches from~FE to~LE and viceversa (see Appendix~\ref{app:FLE}). 

 The version of Algorithm~\ref{alg:BFGS_FD} equipped with a surrogate model is given in Algorithm~\ref{alg:FLE-S} and is denoted as~FLE-S, where the~`S' stands for surrogate. FLE-S enables the use of the surrogate when the cardinality of~$\mathcal{D}_k$ reaches~$\zeta(n+1)(n+2)/2$, where~$\zeta > 0$. The dataset~$\mathcal{D}_k$ includes all points used in the calculation of the~FD gradient up to the current iteration (see Step~1). All the points generated in the~LE step (direct search) are also included in $\mathcal{D}_k$ (see Appendix~\ref{app:FLE}). When employing an interpolation or regression surrogate, among all the points generated in the line search at Steps~8--9, we only include the first one in~$\mathcal{D}_k$ to avoid adding points that are aligned (see Step~11). When using a neural network surrogate, all the points generated in the line search are added to~$\mathcal{D}_k$ (again, see Step~11). At Step~2, we generate an extra point according to a uniform distribution defined over a ball centered at the current iterate and with radius~$0.1$, i.e.,~$B(x_k;0.1)$, and we add such a point to~$\mathcal{D}_k$. Then, at Step~3, when employing an interpolation or regression surrogate, a model can be built by quadratic interpolation or regression or fitting radial basis functions to the current sample set~$Y_k$. When using the neural network surrogate, all the points in~$\mathcal{D}_k$ can be used for training the model.

\begin{algorithm}[H]
\caption{FLE-S (only showing the FD-BFGS step with surrogate)}\label{alg:FLE-S}
\begin{algorithmic}[1]
    %\scriptsize
    \medskip
    \item[] {\bf Input:} Iterates $x_{k-1}$ and $x_{k}$, $\varepsilon > 0$, $\zeta > 0$, backtracking parameters $\bar{\beta} > 0$ and $\tau \in (0,1)$, dataset~$\mathcal{D}_k$. 
  \smallskip 
  \item {\bf If} $|\mathcal{D}_k| < \zeta (n+1)(n+2)/2$, compute the~FD gradient~$g_k$ by using~\eqref{eq:FD}, and add all points used in the calculation of the~FD gradient to $\mathcal{D}_k$.
  \smallskip
  \item[] {\bf Else} 
  \item \quad\quad Add a point uniformly generated on~$B(x_k;0.1)$ to~$\mathcal{D}_k$.
  \item \quad\quad Build an interpolation or regression surrogate or train a neural network surrogate.
  \item \quad\quad Set~$g_k$ equal to the gradient of the surrogate.
  \smallskip
  \item Set~$s_k = x_k - x_{k-1}$ and~$y_k = g_k - g_{k-1}$.
  \smallskip
  \item \textbf{If} $s_k^\top y_k \ge \varepsilon \|s_k\| \|y_k\|$, set~$H_k$ according to~\eqref{eq:H_matrix}, \textbf{else} $H_k = H_{k-1}$.
  \smallskip
\item Compute the direction $p_k = - H_k g_k$ and set $\beta = \bar{\beta}$ for the backtracking line-search.
\smallskip
\item {\bf While} \eqref{eq:FE_suff_decr_noapp} is false {\bf do}
\item \quad\quad Set $\beta = \tau \beta$.
\item Set $\beta_k = \beta$, $x_{k+1} = x_k + \beta_k p_k$. 
\item When using an interpolation or regression surrogate, among all the points generated in the line search at Steps~8--9, add the first one to~$\mathcal{D}_k$. When using a neural network surrogate, add all of these points to~$\mathcal{D}_k$.
\medskip
    \item[] {\bf Output:} Iterate $x_{k+1}$.
    \par\bigskip\noindent
    \end{algorithmic}
\end{algorithm}

%%%%%%%%%%%%%%%%%%%%%%%%%%%%%%%%%%%%%%%%%%%%%%%%%%%%%%%%%%%%%%%%%%%%%%%%%%%%%%%%%%%%%%%%%
\subsection{Numerical experiments}\label{subsec:DFO_numerical_exp}
%%%%%%%%%%%%%%%%%%%%%%%%%%%%%%%%%%%%%%%%%%%%%%%%%%%%%%%%%%%%%%%%%%%%%%%%%%%%%%%%%%%%%%%%%

We used both sets of test problems listed in Tables~\ref{Tab:40probs} and~\ref{Tab:negcurv_probs} of Appendix~\ref{app:tables} to compare the default version of the~FLE algorithm against its version equipped with a model. In particular, we compared the results obtained by~FLE with those obtained by~FLE-S when using the natural basis~\eqref{eq:natural_basis}, basis~\eqref{eq:modmod_natural_basis} with Sigmoid activation, basis~\eqref{eq:RBF}, and neural networks with ReLU and SiLU as surrogates. We did not select basis~\eqref{eq:mod_natural_basis} because it has a similar performance to the natural basis~\eqref{eq:natural_basis}, as discussed in Subsection~\ref{subsec:comparison_activation}. When using basis~\eqref{eq:modmod_natural_basis}, we chose Sigmoid due to its slightly superior performance in the numerical results in Subsection~\ref{subsec:comparison_activation}. When using the neural network surrogates, we chose ReLU and SiLU as activation functions because the numerical results in Subsection~\ref{subsec:comparison_NN} show that they are effective in approximating gradients, with~SiLU being the best at capturing curvature information.  

At Step~3 of Algorithm~\ref{alg:FLE-S}, when employing the natural basis~\eqref{eq:natural_basis}, basis~\eqref{eq:modmod_natural_basis}, or basis~\eqref{eq:RBF}, we built a model on the sample set~$Y_k$, which consisted of all the points~$x_i$ from~$\mathcal{D}_k$ in the ball~$B(x_k; 0.1\gamma)$, with~$\gamma = 1.1^j$, where~$j \in \mathbb{Z}$ is the smallest integer such that~$B(x_k; 0.1\gamma)$ contains at least~$(n+1)(n+2)/2$ points. Therefore, the resulting sample set was~$Y_k = \{x_i ~|~ (x_i,f(x_i)) \in \mathcal{D}_k \text{ and } x_i \in B(x_k; 0.1\gamma)\}$. When using the neural network surrogate, all the points in~$\mathcal{D}_k$ were used for training the model. For the details of the neural network architecture and the algorithm used for training, we refer to Subsection~\ref{subsec:activation_functions_numerical_exp}. We trained the network using a learning rate of~$10^{-2}$, which was chosen by performing a grid search over the set $\{10^{-1},10^{-2},10^{-3},10^{-4}\}$. Regarding the training process, we implemented the following rule for our experiments. The first time the surrogate model is called, the training consists of 5 epochs. Then, only 1 epoch is used to avoid an excessive increase in the computational cost. Regarding the value of the parameter~$\zeta$ in Step~1 of Algorithm~\ref{alg:FLE-S}, we set~$\zeta = 1$ when employing the natural basis~\eqref{eq:natural_basis}, basis~\eqref{eq:modmod_natural_basis}, or basis~\eqref{eq:RBF}, and~$\zeta=0.2$ when using the neural network surrogate. Note that the choice of~$\zeta = 1$ for the interpolation/regression surrogate implies that~FLE-S begins constructing the model once there are enough points to perform at least a determined quadratic interpolation.

\begin{figure}
    \centering
          \includegraphics[scale=0.40]{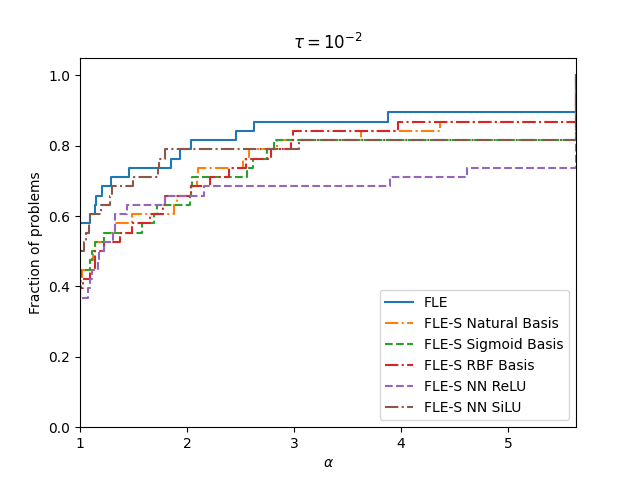} 
          \includegraphics[scale=0.40]{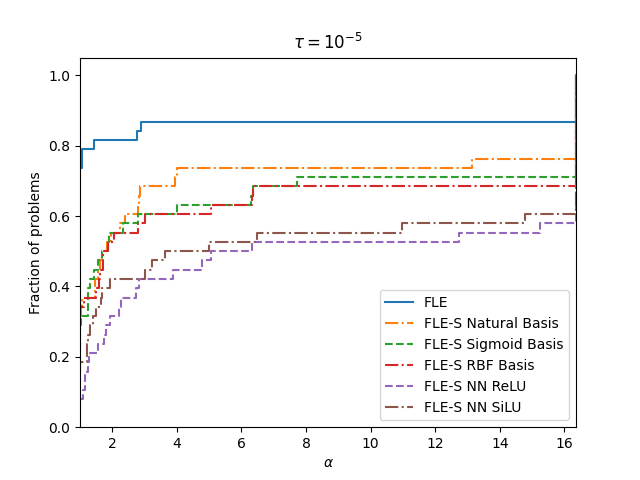} 
          \includegraphics[scale=0.40]{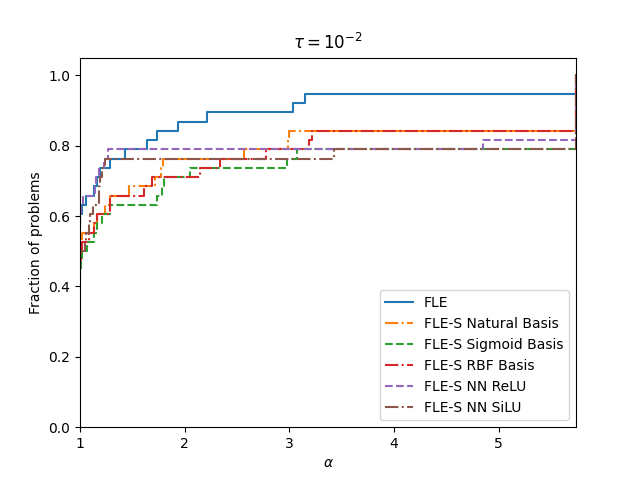} 
          \includegraphics[scale=0.40]{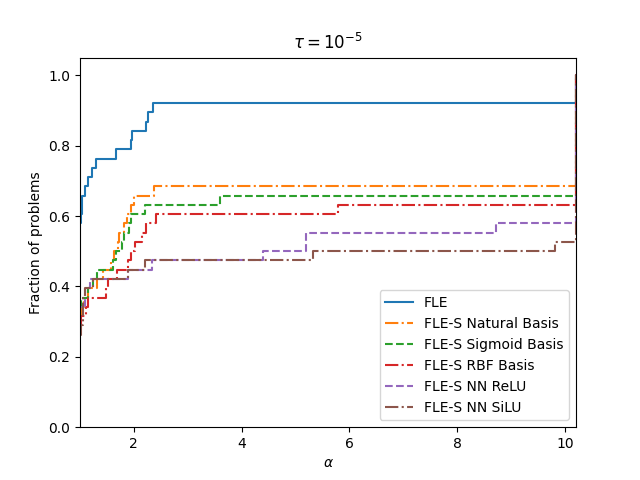} 
          \includegraphics[scale=0.40]{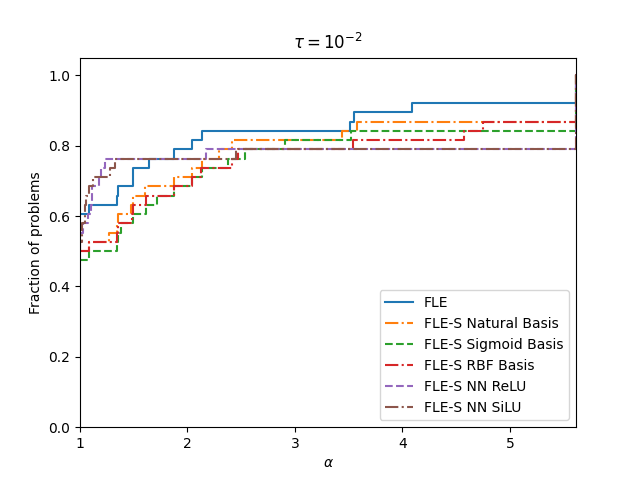} 
          \includegraphics[scale=0.40]{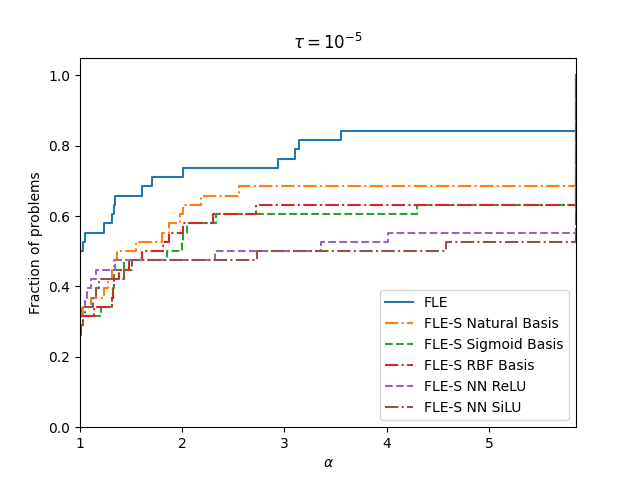}
        \caption{Performance profiles for $n = 20$ (first row), $n = 40$ (second row), and $n = 60$ (third row) with $\tau = 10^{-2}$ and $\tau = 10^{-5}$ on the set of~38 problems listed in Table~\ref{Tab:40probs} of Appendix~\ref{app:tables}. }\label{fig:PP_40probs}
    \end{figure}

\begin{figure}
    \centering
          \includegraphics[scale=0.40]{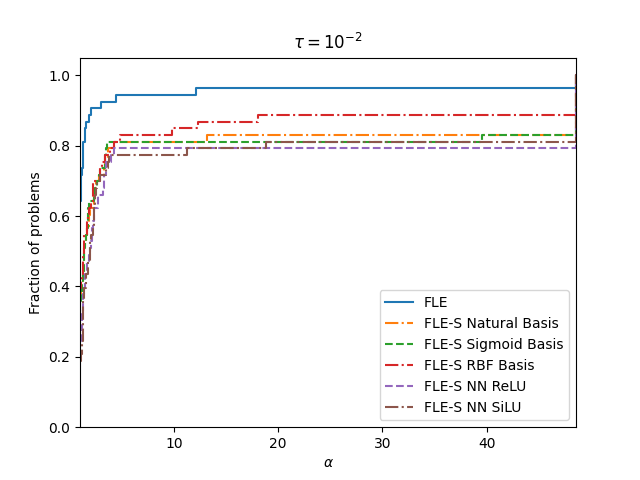} 
          \includegraphics[scale=0.40]{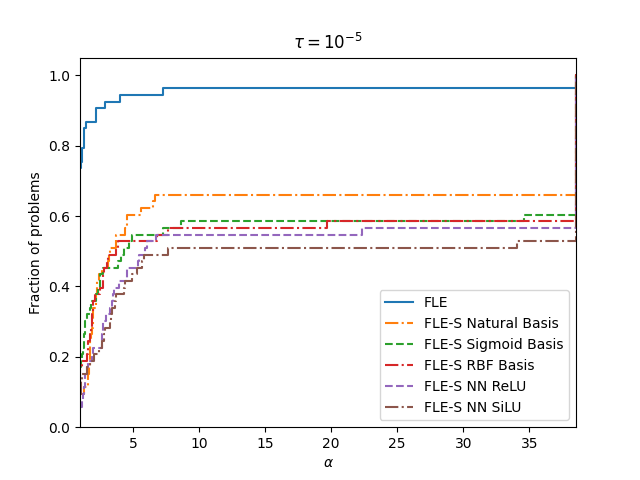}  
        \caption{Performance profiles with $\tau = 10^{-2}$ and $\tau = 10^{-5}$ on the set of~53 problems listed in Table~\ref{Tab:negcurv_probs} of Appendix~\ref{app:tables}.}\label{fig:PP_negcurv_probs}
    \end{figure}

Figures~\ref{fig:PP_40probs} and~\ref{fig:PP_negcurv_probs} show the performance profiles with $\tau = 10^{-2}$ and $\tau = 10^{-5}$ on the sets of~38 and~53 problems, respectively. For each test problem~$p \in \mathcal{P}$, the performance measure~$t_{p,s}$ is given by the number of function evaluations required by each solver~$s \in \mathcal{S}$ to achieve a point~$x$ that satisfies the following convergence test
\[
f(x_0) - f(x) ~\ge~ (1 - \tau) (f(x_0) - f_L),
\]
where~$x_0$ is the starting point for the problem~$p$ and $f_L$ is the smallest value of~$f$ obtained by any solver in~$\mathcal{S}$. Although~FLE requires~$n$ function evaluations more than~FLE-S at each iteration when the model is enabled, FLE-S is not able to outperform~FLE regardless of the type of surrogate used. Note that on the set of~38 problems, when the size of the problems becomes larger and~$\tau=10^{-2}$, FLE-S with the neural network surrogate tends to be more efficient than~FLE, but this is not enough to achieve the same performance as~FLE in terms of robustness. Moreover, when using~$\tau=10^{-5}$, the performance of such an~FLE-S significantly deteriorates both in terms of efficiency and robustness. 
We point out that using a more complex architecture for the neural network surrogate would require more training points to achieve accurate approximations, and so the neural network surrogate would still not be competitive against the default FLE.

%%%%%%%%%%%%%%%%%%%%%%%%%%%%%%%%%%%%%%%%%%%%%%%%%%%%%%%%%%%%%%%%%%%%%%%%%%%%%%%%%%%%%%%%%
\section{Concluding remarks}\label{sec:conclusions}
%%%%%%%%%%%%%%%%%%%%%%%%%%%%%%%%%%%%%%%%%%%%%%%%%%%%%%%%%%%%%%%%%%%%%%%%%%%%%%%%%%%%%%%%%

In this paper, we assessed the use of neural networks as surrogate models to approximate and minimize objective functions in optimization problems. To conduct this analysis, we considered two sets of popular nonlinear optimization test problems. The first set consists of~38 problems with different features (non-linearity, non-convexity, partial
separability), while the second set consists of~53 problems whose objective functions have been observed to have negative curvature. In Section~\ref{sec:activation_functions}, we used performance profiles to compare the performance of neural networks equipped with ReLU, ELU, SiLU, Sigmoid, and Tanh in approximating the objective functions of the test problems in the two sets on training and testing datasets. We found that~SiLU is the best activation function on both sets of problems, ensuring superior efficiency and robustness on the testing dataset. 

In Section~\ref{sec:basis_functions}, we compared the function values, gradients, and Hessians of the objective functions in the two sets of test problems with those of surrogate models obtained through interpolation or regression and neural networks. The numerical results of Subsection~\ref{subsec:comparison_activation} demonstrate that the composition of an activation function with the natural basis can lead to better function value and gradient approximations than the natural basis and basis~\eqref{eq:RBF} when applying quadratic regression with basis~\eqref{eq:modmod_natural_basis} (which does not contain the cross-terms~$x_i x_j$). This suggests that an activation function can waive the necessity of including cross terms in the natural basis. Quadratic interpolation with basis~\eqref{eq:mod_natural_basis} does not lead to any significant improvement compared to the natural basis. On problems with negative curvature, all the bases provide similar performance in terms of function value, gradient, and Hessian approximations. The numerical results of Subsection~\ref{subsec:comparison_NN} show that a neural network surrogate provides better function value and gradient approximations when ReLU or its variants are used as activation functions, while the Hessian approximations are highly inaccurate regardless of the activation function used, with the exception of SiLU. 

In Section~\ref{sec:optimizing_without_derivatives}, we aimed to enhance the performance of the~FLE method by using a surrogate model to approximate the objective function gradient in the~FD-BFGS update step. This approach allows one to avoid calculating a~FD gradient, which requires~$n$ additional function evaluations at each iteration. 
As a surrogate model, we tested the interpolation/regression surrogates obtained from the natural basis~\eqref{eq:natural_basis}, basis~\eqref{eq:modmod_natural_basis}, and basis~\eqref{eq:RBF}, and a neural network surrogate. None of the resulting algorithms was able to significantly enhance the performance of the default~FLE method. The neural network surrogate was observed to be less robust than the interpolation/regression surrogates, although it exhibited higher efficiency when dealing with larger problem sizes. We point out that we also conducted experiments to test the use of the surrogates as replacements for the objective function in the line search at Steps~8--9 of Algorithm~\ref{alg:FLE-S} and the addition of a search step where the surrogate is minimized to determine a better iterate. However, these experiments did not result in a good performance for the algorithms equipped with surrogate models and, therefore, we decided to omit the corresponding numerical results from the paper.

%%%%%%%%%%%%%%%%%%%%%%%%%%%%%%%%%%%%%%

\section*{Acknowledgments}
This work is partially supported by the U.S. Air Force Office of Scientific Research (AFOSR) award FA9550-23-1-0217.

%\section*{Declarations}
%\paragraph{Conflict of interest} The authors have no relevant financial or non-financial interests to disclose. 
%
%\section*{Data availability}
%The authors confirm that the data supporting the findings of this study are available within
%the article.

%\bibliography{ref-BSG,ref-idfo,ref-NN}

\appendix

\section{Appendix: Test Problems from CUTEst}\label{app:tables}

Tables~\ref{Tab:40probs} and~\ref{Tab:negcurv_probs} below list the two sets of nonlinear unconstrained optimization problems from the CUTEst library~\cite{NGould_DOrban_PToint_2015} used throughout the paper and described in Subsection~\ref{subsec:activation_functions_numerical_exp}.

\begin{table}[ht]
\centering
\begin{tabular}{l|l|l|l}
\hline
ARGLINA & ARGTRIGLS & ARWHEAD & BDEXP \\
BOXPOWER & BROWNAL & COSINE & CURLY10 \\
DQRTIC & DIXON3DQ & ENGVAL1 & EXTROSNB \\ FLETBV3M & FLETCBV3 & FLETCHBV & FLETCHCR \\ FREUROTH & INDEFM & MANCINO & MOREBV \\NONCVXU2 & NONCVXUN & NONDIA & NONDQUAR \\ PENALTY2 & POWER & QING & QUARTC \\ SENSORS & SINQUAD & SCURLY10 & SCURLY20 \\ SPARSINE & SPARSQUR & SSBRYBND & TRIDIA \\ TRIGON1 & TOINTGSS & \\
\hline
\end{tabular}
\caption{Names of the~38~CUTEst problems with user-defined dimensions in the first set.}\label{Tab:40probs}
\end{table}

\begin{table}[ht]
\centering
\begin{tabular}{l|c|l|c|l|c}
\hline
\textbf{Name} & \textbf{Dimension} & \textbf{Name} & \textbf{Dimension} & \textbf{Name} & \textbf{Dimension} \\
\hline
ALLINITU & 4 & BARD & 3 & BIGGS6 & 6 \\
BOX3 & 3 & BROYDN7D & 10 & BRYBND & 10 \\
CUBE & 2 & DENSCHND & 3 & DENSCHNE & 3 \\
DIXMAANA1 & 15 & DIXMAANB & 15 & DIXMAANC & 15 \\
DIXMAAND & 15 & DIXMAANE1 & 15 & DIXMAANF & 15 \\
DIXMAANG & 15 & DIXMAANH & 15 & DIXMAANI1 & 15 \\
DIXMAANJ & 15 & DIXMAANK & 15 & DIXMAANL & 15 \\
ENGVAL2 & 3 & ERRINROS & 25 & EXPFIT & 2 \\
FMINSURF & 16 & GROWTHLS & 3 & GULF & 3 \\
HAIRY & 2 & HATFLDD & 3 & HATFLDE & 3 \\
HEART6LS & 6 & HEART8LS & 8 & HELIX & 3 \\
HIELOW & 3 & HIMMELBB & 2 & HIMMELBG & 2 \\
HUMPS & 2 & KOWOSB & 4 & LOGHAIRY & 2 \\
MARATOSB & 2 & MEYER3 & 3 & MSQRTALS & 4 \\
MSQRTBLS & 9 & OSBORNEA & 5 & OSBORNEB & 11\\
PENALTY3 & 50 & SNAIL & 2 & SPMSRTLS & 28 \\
STRATEC & 10 & VIBRBEAM & 8 & WATSON & 12\\
WOODS & 4 & YFITU & 3 & & \\
\hline
\end{tabular}
\caption{Names and corresponding dimensions of the~53~CUTEst problems in the second set.}\label{Tab:negcurv_probs}
\end{table}

%%%%%%%%%%%%%%%%%%%%%%%%%%%%%%%%%%%%%%%%%%%%%%%%

\section{Appendix: Full-Low Evaluation Methods}\label{app:FLE}

In DFO, two main algorithmic paradigms are employed when designing numerical algorithms, namely, \textit{directional} and \textit{model-based} approaches~\cite{ARConn_KScheinberg_LNVicente_2009}. In directional algorithms, a set of directions is generated to determine a point that guarantees a (possibly sufficient) decrease condition of the objective function among a set of polling points, which are obtained from the current iterate by considering certain stepsizes along these directions. When only objective function values are used, without approximating the gradients or constructing models, the resulting algorithms are referred to as directional direct-search methods~\cite{CAudet_JEDennis_2006,AJBooker_etal_1998,SGratton_etal_2015,TGKolda_RMLewis_VTorczon_2003,VTorczon_1997,TGiovannelli_etal_2022}. Such algorithms are able to converge even when applied to problems with a non-smooth objective function~\cite{CAudet_JEDennis_2006,LNVicente_ALCustodio_2012}. Within this class of methods, one can consider deterministic variants~\cite{TGKolda_RMLewis_VTorczon_2003,VTorczon_1997} or probabilistic ones~\cite{SGratton_etal_2015}. Deterministic variants rely on positive spanning sets of vectors, where at least one of them is a descent direction. Probabilistic variants use randomly generated directions that are probabilistically descent. To determine the value of the stepsize, DFO algorithms can use line searches along a prespecified set of directions~\cite{SLucidi_MSciandrone_2002,GFAsano_GLiuzzi_etal_2014} or along directions that are obtained by applying a~FD scheme to approximate the gradient of the objective function~\cite{ASBerahas_etal_2019,ASBerahas_LCao_KScheinberg_2021}. In noisy settings, such approximations can be unreliable unless careful estimation of the noise is considered~\cite{JJMore_SMWild_2011,JJMore_SMWild_2012}. 

In model-based algorithms, a model of the objective function is built by using function values at previous iterates or at points randomly sampled~\cite{ARConn_PhLToint_1996,ARConn_KScheinberg_LNVicente_2009b,GFasano_JLMorales_JNocedal_2008,MJDPowell_2004b,DWinfield_1973,GLiuzzi_etal_2019}. Such methods rely on interpolation or regression techniques and use basis functions such
as quadratic polynomials or radial basis functions. The resulting models can be used as a local approximation of the objective function to capture its curvature. A drawback of model-based algorithms is that their performance gets worse as the number of variables increases because of the dense linear algebra of the interpolation and the ill-conditioning of the sample sets~\cite{ARConn_KScheinberg_LNVicente_2006}. Moreover, noisy and non-smooth settings remain a challenge for such algorithms.

The~FLE method introduced in Section~\ref{sec:optimizing_without_derivatives} combines the benefits of both directional and model-based algorithms and exhibited superior overall performance compared to interpolation-based methods and combinations of these techniques with direct search~\cite{ASBerahas_OSohab_LNVicente_2023}. In the~FLE method, the first iteration is of~FE type and is given by Algorithm~\ref{alg:BFGS_FD}. An~FE iteration is no longer deemed successful when the stepsize~$\beta_k$ becomes too small, i.e.,
\begin{equation}\label{eq:FE_stepsize}
    \beta ~<~ \lambda \rho(\alpha_k),
\end{equation}
where~$\lambda > 0$ and $\rho(\alpha_k)$ is a forcing function depending on the stepsize used in~LE iterations, denoted as~$\alpha_k$. In such a case, one switches to an~LE iteration, which consists of a direct-search step.  

The schema of an LE iteration is reported in Algorithm~\ref{alg:LE}. At each iteration, the stepsize~$\alpha_k$ is required to satisfy the following sufficient decrease condition 
\begin{equation}\label{eq:LE_suff_decr}
    f(x_k + \alpha_k d_k) ~\le~ f(x_k) - \rho(\alpha_k),
\end{equation}
where~$\rho(\alpha_k)$ is the same forcing function used in~\eqref{eq:FE_stepsize} and~$d_k \in D_k$ is a polling direction. In the practical~FLE method proposed in~\cite{ASBerahas_OSohab_LNVicente_2023}, the set of polling directions~$D_k$ is given by a direction~$d \in \mathbb{R}^n$ uniformly generated on the unit sphere of~$\mathbb{R}^n$ and its negative, i.e., $D = [d,-d]$. An~LE iteration is deemed successful when a stepsize satisfying~\eqref{eq:LE_suff_decr} is found. One can switch from an~LE iteration to an~FE iteration when the number of unsuccessful consecutive~LE iterations becomes greater than the number of line-search backtracks done in the previous~FE iteration.  

	 \begin{algorithm}[H]
	\caption{\texttt{Low-Eval Iteration}: Direct Search}\label{alg:LE}
	\begin{algorithmic}[1]
		%\scriptsize
		\medskip
		\item[] {\bf Input:} Iterate $x_k$ and stepsize $\alpha_k$. Direct-search parameters $\lambda \ge 1$ and $\theta \in (0,1)$. \smallskip 
  \item Generate a finite set $D_k$ of non-zero polling directions.
  \item \textbf{If} \eqref{eq:LE_suff_decr} is true for some $d_k \in D_k$, set $x_{k+1} = x_k + \alpha_k d_k$ and $\alpha_{k+1} = \lambda \alpha_k$.
  \item \textbf{Else} set $x_{k+1} = x_k$ and $\alpha_{k+1} = \theta \alpha_k$.
		\item Decide if $t_{k+1} = \texttt{Low-Eval}$ or if $t_{k+1} = \texttt{Full-Eval}$.
\medskip
        \item[] {\bf Output:} $t_{k+1}$, $x_{k+1}$, and $\alpha_{k+1}$.
		\par\bigskip\noindent
    	\end{algorithmic}
    \end{algorithm}

For the values of the parameters of the~FLE method used in the experiments reported in Section~\ref{sec:optimizing_without_derivatives}, we refer to the practical~FLE algorithm in~\cite{ASBerahas_OSohab_LNVicente_2023}.
    
\end{document}